\documentclass[10pt,twocolumn,letterpaper]{article}

\usepackage{cvpr}              

\usepackage{times}
\usepackage{epsfig}

\usepackage{graphicx}
\usepackage{amsmath}
\usepackage{amssymb}
\usepackage{booktabs}

\usepackage{amsfonts}       
\usepackage{nicefrac}       
\usepackage{microtype}      
\usepackage[dvipsnames]{xcolor}         
\usepackage{bm}             
\usepackage{amsthm}         
\usepackage{xspace}         
\usepackage{adjustbox}      
\usepackage{multirow}       
\usepackage{pifont}         
\usepackage{caption}        
\usepackage[ruled,vlined]{algorithm2e}  
\newcommand{\boldx}{\bm{x}}                                 
\newcommand{\boldy}{\bm{y}}                                 
\newcommand{\boldw}{\bm{w}}                                 
\newcommand{\estscoreguide}{\bm{s}_{\bm{\theta}}}           
\newcommand{\estscoreori}{\bm{s}_{\bm{\phi}}}               
\newcommand{\estscoreoriabstract}{\bm{s}}                   
\usepackage{eqparbox}                                       
\newcommand{\up}{$\uparrow$}                                
\newcommand{\down}{$\downarrow$}                            
\newcommand{\psnr}{PSNR}                                    
\newcommand{\cyclediff}[0]{CycleDiffusion\xspace}           

\usepackage{listings}                                       

\usepackage{inconsolata}

\definecolor{mdgreen}{rgb}{0.05,0.6,0.05}
\definecolor{mdblue}{rgb}{0,0,0.7}
\definecolor{dkblue}{rgb}{0,0,0.5}
\definecolor{dkgray}{rgb}{0.3,0.3,0.3}
\definecolor{slate}{rgb}{0.25,0.25,0.4}
\definecolor{gray}{rgb}{0.5,0.5,0.5}
\definecolor{ltgray}{rgb}{0.7,0.7,0.7}
\definecolor{lavender}{rgb}{0.65,0.55,1.0}

\definecolor{mypurple}{RGB}{111,61,121}
\definecolor{myred}{RGB}{181,68,106}

\definecolor{hanblue}{rgb}{0.27, 0.42, 0.81}                
\newcommand{\guidedcolor}[1]{\textcolor{hanblue}{#1}}       
\newcommand{\contrastcolor}[1]{\textcolor{purple}{#1}}         

\newcommand{\anonymoustext}[1]{}                          
\newcommand{\acceptedtext}[1]{#1}                             
\newcommand{\opensource}{\url{https://github.com/humansensinglab/contrastive-guidance}}   

\newcommand{\dalle}[0]{DALL$\cdot$E\xspace}  
\newcommand{\ourmethod}[0]{\textit{Contrastive Guidance}\xspace}  
\newcommand{\titleourmethod}[0]{Contrastive Guidance\xspace}  
\newcommand{\cfg}[0]{CFG\xspace}  

\definecolor{cvprblue}{rgb}{0.21,0.49,0.74}
\usepackage[pagebackref,breaklinks,colorlinks,citecolor=cvprblue]{hyperref}

\def\paperID{*****} 
\def\confName{CVPR}
\def\confYear{2024}

\title{Contrastive Prompts Improve Disentanglement in Text-to-Image Diffusion Models}

\author{
{Chen Henry Wu, \ Fernando De la Torre} \\
Carnegie Mellon University, Pittsburgh, PA \\
{\tt \{chenwu2,ftorre\}@cs.cmu.edu}
}

\begin{document}

\twocolumn[{%
\renewcommand\twocolumn[1][]{#1}%
\maketitle

\begin{center}
    \centering
    \includegraphics[width=0.95\linewidth]{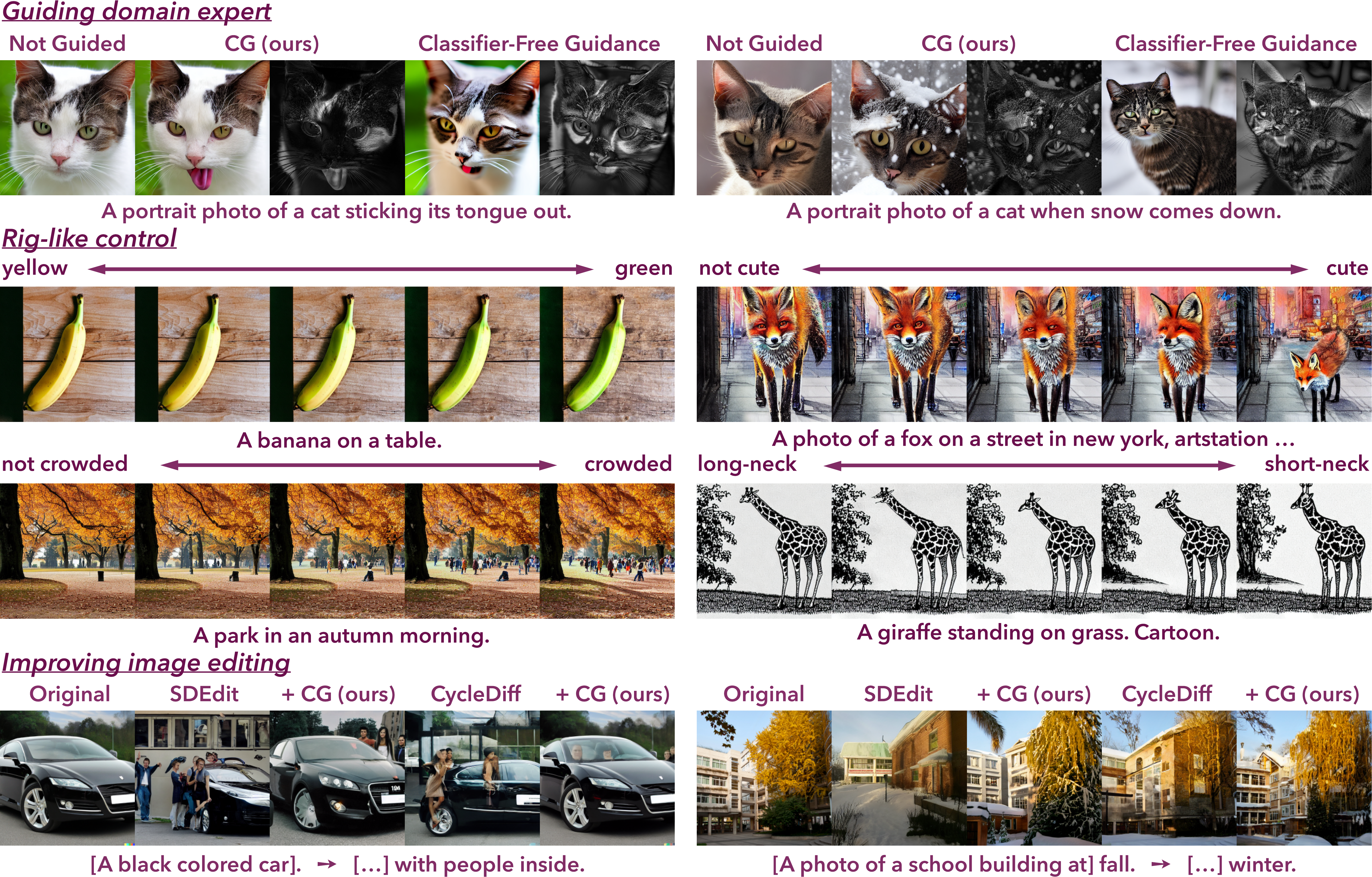}
    \captionof{figure}{This paper explores \textit{disentangled control} for text-to-image diffusion models. We show that using a pair of \textit{contrastive prompts} can (1) guide domain experts (e.g., trained on an object class) with text, (2) enable rig-like control for text-to-image diffusion models (similar to steering a StyleGAN), and (3) improve text-to-image diffusion-based image editing methods.}%
    \label{fig:teaser}
\end{center}
}]

\maketitle


\begin{abstract} 
Text-to-image diffusion models have achieved remarkable performance in image synthesis, while the text interface does not always provide fine-grained control over certain image factors. For instance, changing a single token in the text can have unintended effects on the image. This paper shows a simple modification of classifier-free guidance can help disentangle image factors in text-to-image models. The key idea of our method, \textbf{\ourmethod}, is to characterize an intended factor with two prompts that differ in minimal tokens: the positive prompt describes the image to be synthesized, and the baseline prompt serves as a ``baseline'' that disentangles other factors. \ourmethod is a general method we illustrate whose benefits in three scenarios: (1) to guide domain-specific diffusion models trained on an object class, (2) to gain continuous, rig-like controls for text-to-image generation, and (3) to improve the performance of zero-shot image editors. Code available at: \opensource{}
\end{abstract}

\section{Introduction}

Many recent methods for text-to-image synthesis are built upon diffusion models trained on large-scale datasets (e.g., GLIDE \cite{Nichol2021GLIDETP}, \dalle 2 \cite{Ramesh2022HierarchicalTI}, Imagen \cite{Saharia2022PhotorealisticTD}, and Stable Diffusion \cite{Rombach2021ldm}).
Besides generating images with unprecedented visual quality, these models also show remarkable generalization ability to novel texts with complex compositions of concepts.

Prompt engineering and guidance \cite{dhariwal2021diffusion,ho2021classifierfree} are two (overlapping) directions for steering the outputs of these text-to-image models. Prompt engineering focuses on how to design better text as input to the model, while guidance focuses on how the image distribution can be modified by text. However, prompt engineering and previous guidance methods \cite{dhariwal2021diffusion,ho2021classifierfree,Nichol2021GLIDETP,Liu2022CompositionalVG} still face challenges in fine-grained control over image factors. They can affect many image factors, not only what is intended. Thus, in this paper, we raise the following question. \textit{How can we steer the outputs of text-to-image diffusion models over certain image factors, without affecting other factors? } 

In this paper, we show that a simple modification of classifier-free guidance can help disentangle image factors in text-to-image diffusion models. Our key idea is to instantiate an intended image factor with two text prompts that differ only in several tokens: the \textit{positive prompt} describes the image to be synthesized, while the \textit{baseline prompt} serves as a ``baseline'' that disentangles other factors. The difference between the two score functions conditioned on the two prompts is used as guidance. For instance, suppose we already have a diffusion model for high-quality cat images. To synthesize a cat on the beach, one can set the positive prompt as \textit{a portrait photo of a cat on the beach}, and the baseline prompt as \textit{a portrait photo of a cat}. The baseline prompt helps the model focus on \textit{on the beach}, preventing it from contributing to what \textit{a portrait photo of cat} looks like. 

Besides the above intuition, we also formalize \ourmethod in a probabilistic framework. Specifically, by applying the Bayes rule, we show that the positive and baseline prompts together build a binary generative classifier over the noisy images. This binary generative classifier can be used to guide the denoising process following classifier guidance \cite{dhariwal2021diffusion}. Note that, in classifier guidance, the score function of the classifier is computed by backpropagating through the classifier. Different from them, we show that our generative classifier has a gradient-free solution that corresponds to the score functions of the two prompts, and \ourmethod is proportional to this solution up to a constant.

Since our derivation is based on the unified SDE/ODE formulation \cite{0011SKKEP21}, \ourmethod can be plugged into different diffusion models and samplers/schedulers in various scenarios: 
\begin{itemize}
    \item Guiding Domain Expert (Figure~\ref{fig:teaser} top): Given a \textit{domain expert} trained on high-quality cat faces, how can we combine the domain knowledge of the domain expert with the generalization ability of the text-to-image model? We show that compared with classifier-free guidance \cite{ho2021classifierfree}, our \ourmethod can better disentangle the domain knowledge and text condition. 
    \item Continuous, Rig-like Control (Figure~\ref{fig:teaser} middle): How could we gain continuous, rig-like control over the color while maintaining others? It helps synthesize the same object with interpolated color. 
    \item Improving Image Editing (Figure~\ref{fig:teaser} bottom): Previous works have used text-to-image diffusion models for image editing \cite{Hertz2022PrompttoPromptIE,unify,meng2022sdedit,Kawar2022ImagicTR}. Disentanglement is the key to this application since the user only wants the intended edit (e.g., adding people) without altering other factors (e.g., the car and the background). 
\end{itemize}
Our experiments verify these applications. Furthermore, our analysis demonstrates that our \ourmethod is more disentangled than classifier-free guidance \cite{ho2021classifierfree}.

\section{Related Work}
\label{sec:related-work}

\paragraph{Text-to-image diffusion model: } 
\ Recent years have witnessed unprecedented progress in text-to-image synthesis, driven by large generative models such as diffusion models \cite{HoJA20,SongE19,0011SKKEP21,dhariwal2021diffusion} (e.g., GLIDE \cite{Nichol2021GLIDETP}, \dalle 2 \cite{Ramesh2022HierarchicalTI}, Imagen \cite{Saharia2022PhotorealisticTD}, and Stable Diffusion \cite{Rombach2021ldm}) and VQ Transformers \cite{Esser2021TamingTF} (e.g., \dalle \cite{dalle}, CogView \cite{Ding2021CogViewMT,Ding2022CogView2FA}, and Parti \cite{Yu2022ScalingAM}). This paper studies how to extract disentangled image factors from text-to-image diffusion models.

\paragraph{Guidance for diffusion models. }
Guidance methods modify the output distribution of pre-trained diffusion models, based on additional inputs such as class labels \cite{dhariwal2021diffusion}, text \cite{Nichol2021GLIDETP}, and corrupted images \cite{Kawar2022DenoisingDR,Lugmayr2022RePaintIU}.  The first guidance method is classifier guidance \cite{dhariwal2021diffusion}, for which a class classifier is finetuned on noisy images. Similarly, CLIP guidance \cite{Nichol2021GLIDETP,Liu2021MoreCF} finetunes a CLIP model \cite{Radford2021LearningTV} to support text input. To avoid finetuning classifiers or CLIP, classifier-free guidance (CFG) \cite{ho2021classifierfree} jointly trains a conditional and an unconditional diffusion model and combines their score estimates, and CFG has become the default for text-to-image tasks \cite{Nichol2021GLIDETP,Rombach2021ldm}. To compose multiple texts, composable diffusion \cite{Liu2022CompositionalVG} combines score estimates with different text inputs. Besides user-specified conditions, several works showed that even guidance based on model outputs \cite{Chen2022AnalogBG} or representations \cite{Hong2022ImprovingSQ} can improve the quality of images. In this paper, we explore how to disentangle image factors with texts to gain fine-grained control. 

\paragraph{Image editing with diffusion models. } Recent works have shown that diffusion models are capable of unpaired image-to-image translation \cite{meng2022sdedit,Su2022DualDI,Choi2021ILVRCM,unify}. A more recent trend of works have explored zero-shot image editing with text-to-image diffusion models \cite{unify,Hertz2022PrompttoPromptIE,Kawar2022ImagicTR}. One of the applications of our \ourmethod is to improve the intended edit of some of these zero-shot image editors. 

\paragraph{Guidance for other generative models. } 
Guidance has also been widely studied for GANs \cite{Goodfellow2014GenerativeAN} and autoregressive language models (LMs). For GANs, the conditions for guidance include class labels \cite{Nguyen2017PlugP,nie2021controllable} text prompts \cite{Patashnik2021StyleCLIPTM,promptgen}, corrupted images \cite{Menon2020PULSESP}, and de-biasing \cite{Karakas2022FairStyleDS,promptgen}. Although LMs have very different formulations from generative models for images, similar ideas to ours have been explored. For instance, GeDi \cite{gedi} uses a small LM with two contrastive prompts to guide a large LM, and contrastive decoding \cite{Li2022ContrastiveDO} leverages a small language model to correct mistakes by a large LM. 

\section{Method}
\label{sec:method}

\subsection{Background on Diffusion Models}
Diffusion models sample data via iterative denoising. Suppose noises are added to the image $\boldx$ via $\mathrm{d}\boldx = \bm{f}(\boldx, t)\mathrm{d}t + g(t)\mathrm{d}\boldw$, where $t \in [0, T]$ is the time step, and $\bm{f}$ and $g$ are pre-defined functions that specify how exactly the noises are added. It has been shown that the denoising problem can be formulated by the following SDE \cite{0011SKKEP21}:
\begin{equation}
\label{eq:sde-sampling-score}
    \mathrm{d}\boldx = \Big[\bm{f}(\boldx, t) - g(t)^{2} \nabla_{\boldx} \log p_{t}(\boldx) \Big]\mathrm{d}t + g(t)\mathrm{d}\bar{\boldw},
\end{equation}
where $\nabla_{\boldx} \log p_{t}(\boldx)$ is the \textit{score function}, the core of diffusion models, which provides information about the denoising direction. Besides the SDE-based formulation, previous works have also proposed non-Markovian \cite{song2021denoising,Watson2022LearningFS} and ODE-based formulations \cite{0011SKKEP21,song2021denoising,Salimans2022ProgressiveDF,Liu2022PseudoNM,Lu2022DPMSolverAF,Karras2022ElucidatingTD,Zhang2022FastSO}. Since the true score function is inaccessible, diffusion models learn to approximate it $\nabla_{\boldx} \log p_{t}(\boldx) \approx \guidedcolor{\estscoreoriabstract(\boldx, t)}$. For instance, we can plug it into the SDE and get the denoising process: 
\begin{equation}
\label{eq:sde-sampling}
    \mathrm{d}\boldx = \Big[\bm{f}(\boldx, t) - g(t)^{2} \guidedcolor{\estscoreoriabstract(\boldx, t)} \Big]\mathrm{d}t + g(t)\mathrm{d}\bar{\boldw}.
\end{equation}
This connection between diffusion models and differential equations provides a unified formulation for guided image synthesis. Given a condition $c$, one only need to compute the conditional score function $\nabla_{\boldx} \log p_{t}(\boldx|c)$ and replace $\nabla_{\boldx} \log p_{t}(\boldx)$ with $\nabla_{\boldx} \log p_{t}(\boldx|c)$ in the denoising process. For instance, we can plug it into the above SDE and get the conditional denoising process  \cite{0011SKKEP21}:
\begin{equation}
    \mathrm{d}\boldx = \Big[\bm{f}(\boldx, t) - g(t)^{2} \nabla_{\boldx} \log p_{t}(\boldx|c) \Big]\mathrm{d}t + g(t)\mathrm{d}\bar{\boldw}.
\end{equation}
The true conditional score function $\nabla_{\boldx} \log p_{t}(\boldx|c)$ is also inaccessible, and each guidance method aims to approximate it. For instance, given a text-to-image diffusion model $\estscoreguide$ and a text $\boldy$ as the condition, classifier-free guidance (CFG) \cite{ho2021classifierfree} models the conditional denoising process as
\begin{equation}
\begin{split}
    &\mathrm{d}\boldx = \bigg[\bm{f}(\boldx, t) - g(t)^{2} \Big(\tau_t\estscoreguide(\boldx, t, \boldy) \ - \\
    &\quad\quad\quad\quad\quad\quad (\tau_t - 1)\estscoreguide(\boldx, t, \varnothing) \Big) \bigg]\mathrm{d}t + g(t)\mathrm{d}\bar{\boldw},
\end{split}
\end{equation}
where $\varnothing$ is an empty string and $\tau_t$ is a hyperparameter indicating the guidance strength. It implies that CFG approximates $\nabla_{\boldx} \log p_{t}(\boldx|c) \approx \tau_t\estscoreguide(\boldx, t, \boldy) - (\tau_t - 1)\estscoreguide(\boldx, t, \varnothing)$.

\subsection{\titleourmethod}
\label{subsec:method-diffusion}

We introduce \ourmethod, which uses text-to-image diffusion models to steer certain image factors while not affecting others. Our key idea is to instantiate a condition $c$ with two prompts $(\boldy^{+}, \boldy^{-})$ that differ in a few tokens: the \textit{positive prompt} $\boldy^{+}$ describes the intended image, while the \textit{baseline prompt} $\boldy^{-}$ serves as a ``baseline'' that disentangles other factors.\footnote{Although we use $\boldy^{-}$ to denote the baseline prompt, it does not need to be a ``negative prompt''. That is, the baseline prompt need not describe \textit{what not to generate}. See the following text for an example. } Suppose we have a domain expert that can generate high-quality images of cats. To generate a cat with eyeglasses, one can set the positive prompt $\boldy^{+}$ as \textit{a portrait photo of a cat with eyeglasses}, and the baseline prompt $\boldy^{-}$ as \textit{a portrait photo of a cat without eyeglasses}, or simply, \textit{a portrait photo of a cat}. The baseline prompt helps the model focus on \textit{with eyeglasses}, preventing it from contributing to what \textit{a portrait photo of a cat} looks like. In this example, cats are better modeled by the domain expert. 

\begin{figure}[!t]
\centering
  \includegraphics[width=0.98\linewidth]{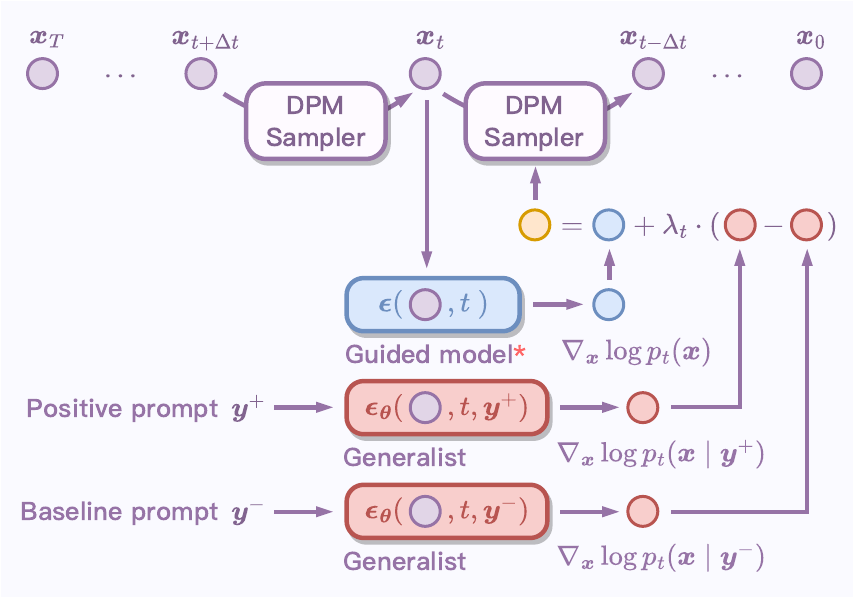}
\caption{\label{fig:method} \ Overview of \ourmethod. Given a model to be guided, \ourmethod use two prompts with minimal differences as guidance signal, which helps disentangle the intended image factors.
\textcolor{red}{$^{*}$}The guided model can be a domain expert, the text-to-image model itself, or an image editing method (Section~\ref{subsec:application-method}).
}
\end{figure}

In the following, we formalize the above intuition. Our goal is to derive $\nabla_{\boldx} \log p_{t}(\boldx|c)$. First, we rewrite $p_{t}(\boldx|c)$ as
\begin{align}
\label{eq:decompose}
    p_{t}(\boldx|c) \propto p_{t}(\boldx) p_{t}(c|\boldx).
\end{align}
We will discuss the model to be guided $p_{t}(\boldx)$ later, which has several variations. The key idea of our formulation is to rewrite the term $p_{t}(c|\boldx)$ as a \textit{generative classifier} using the Bayes rule. It is a binary classifier that infers whether the noisy $\boldx$ is from the positive prompt $\boldy^{+}$ or the baseline prompt $\boldy^{-}$. Formally, we define the generative classifier as
\begin{align}
\label{eq:generative-classifier}
    p_{t}(c|\boldx) := \frac{p(\boldy^{+})p_{t}(\boldx|\boldy^{+})^{\gamma_t}}{p(\boldy^{+})p_{t}(\boldx|\boldy^{+})^{\gamma_t} + p(\boldy^{-})p_{t}(\boldx|\boldy^{-})^{\gamma_t}},
\end{align}
where the temperature $\gamma_t$ controls the sharpness of the distribution. In the following, we will combine Eq.~(\ref{eq:decompose}) and Eq.~(\ref{eq:generative-classifier}), using simplified notations: $p^{+} := p(\boldy^{+})$, $p^{-} := p(\boldy^{-})$, $p_{t}^{+}(\boldx) := p_{t}(\boldx | \boldy^{+})$, and $p_{t}^{-}(\boldx) := p_{t}(\boldx | \boldy^{-})$. Using the \textit{log-derivative trick}, we have (see details in Appendix~\ref{subapp:derivation-score}):
\begin{align}
\begin{split}
    &\quad \ \nabla_{\boldx} \log p_{t}(\boldx|c) \\
    &= \nabla_{\boldx} \log \Big( p_{t}(\boldx) \frac{p^{+}p_{t}^{+}(\boldx)^{\gamma_t}}{p^{+}p_{t}^{+}(\boldx)^{\gamma_t} + p^{-}p_{t}^{-}(\boldx)^{\gamma_t}} \Big)
\end{split}\\
\begin{split}
    &= \nabla_{\boldx} \log p_{t}(\boldx) + \nabla_{\boldx} \log p_{t}^{+}(\boldx)^{\gamma_t} \\
    &\quad\quad\quad\quad\quad\quad - \frac{\nabla_{\boldx} \Big( p^{+}p_{t}^{+}(\boldx)^{\gamma_t} + p^{-}p_{t}^{-}(\boldx)^{\gamma_t} \Big)}{p^{+}p_{t}^{+}(\boldx)^{\gamma_t} + p^{-}p_{t}^{-}(\boldx)^{\gamma_t}}
\end{split}\\
\begin{split}
    \label{eq:contain-lambda}
    &=\nabla_{\boldx} \log p_{t}(\boldx) + \underbrace{\frac{\gamma_t \cdot p^{-}p_{t}^{-}(\boldx)^{\gamma_t}}{p^{+}p_{t}^{+}(\boldx)^{\gamma_t} + p^{-}p_{t}^{-}(\boldx)^{\gamma_t}}}_{\lambda_t} \cdot \\ 
    &\quad\quad\quad\quad\quad\quad \Big(\nabla_{\boldx} \log p_{t}^{+}(\boldx) - \nabla_{\boldx} \log p_{t}^{-}(\boldx) \Big)
\end{split}\\
\begin{split}
    \label{eq:lambda-simplification}
    &=\nabla_{\boldx} \log p_{t}(\boldx) + \lambda_t \Big(\nabla_{\boldx} \log p_{t}^{+}(\boldx) \\
    &\quad\quad\quad\quad\quad\quad\quad\quad\quad\quad\quad - \nabla_{\boldx} \log p_{t}^{-}(\boldx) \Big) 
\end{split}\\
    \label{eq:nn-approximation}
    &\approx \guidedcolor{\estscoreoriabstract(\boldx, t)} + \contrastcolor{\lambda_t \big( \estscoreguide(\boldx, t, \boldy^{+}) - \estscoreguide(\boldx, t, \boldy^{-}) \big)}, 
\end{align}
where $\estscoreguide$ is the text-to-image diffusion model. 
We will introduce several choices of the score estimate $\guidedcolor{\estscoreoriabstract(\boldx, t)}$ in Section~\ref{subsec:application-method}, e.g., domain experts, text-to-image models themselves, and image editors. Please see Figure~\ref{fig:method} for an illustration of how the term is computed with pretrained text-to-image diffusion models and the two prompt conditions. 

\paragraph{(Not) estimating $\lambda_t$} 
A remaining question is how to estimate the coefficient $\lambda_t$. Since $\gamma_t$, $p(\boldy^{+})$ and $p(\boldy^{-})$ do not depend on $\boldx$, the only terms to estimate are $p_{t}(\boldx | \boldy^{+})$ and $p_{t}(\boldx | \boldy^{-})$. We can leverage the fact that each SDE has an ODE with the same marginal density at each time $t$ \cite{Maoutsa2020InteractingPS,0011SKKEP21}. Estimating the density of ODEs is tractable by calling an ODE solver \cite{Chen2018NeuralOD,Grathwohl2019FFJORDFC} from $t$ to $T$ (see Appendix~\ref{subapp:estimate-lambda}). However, this requires calling the ODE solver at all $t$'s, resulting in the total complexity of $\mathcal{O}(N^2)$ where $N$ is the number of discretization steps. To avoid this, this paper assumes that the temperature $\gamma_t$ is adaptive such that $\lambda_t$ does not depend on $\boldx$, giving us $\mathcal{O}(N)$ complexity. Further, our experiments show that a fixed $\lambda_t$ for all time $t$ provides reasonable results. 

\paragraph{Comparison to CFG and negative prompts } 
It is worth noting that although the derivation of the ultimate solution is complex, the final expression in Eq.~(\ref{eq:nn-approximation}) is simple. In terms of the final expression, our approach can be seen as an extension upon classifier-free guidance (CFG) \cite{ho2021classifierfree}, where we employ a non-trivial baseline prompt in place of an empty string. Similarly, negative prompts have become a common practice in most diffusion model applications. Typical negative prompts are phrases describing unwanted outputs such as ``low quality'' and ``NSFW''. 
We would like to highlight the idea of ``specifying the image factor with a contrastive baseline prompt'', which is why our method achieves better disentanglement than CFG (Section~\ref{sec:visualization}). 
In the following sections, we will also demonstrate how this formulation leads to new applications, such as merging a text-to-image diffusion model and a domain-specific diffusion model, rig-like controls, and image editing. 

\subsection{Applications}
\label{subsec:application-method}

\paragraph{Guiding Domain Experts} Text-to-image models sometimes underperform domain-specific models in highly specialized domains (e.g., faces). To combine the best of both worlds, we can use \ourmethod to guide domain experts. Suppose we have a domain expert $\guidedcolor{\estscoreori(\boldx, t)}$ for high-quality cat faces. To generate a cat with eyeglasses, one can set $\boldy^{+}$ as \textit{a portrait photo of a cat with eyeglasses}, and $\boldy^{-}$ as \textit{a portrait photo of a cat without eyeglasses}, or simply, \textit{a portrait photo of a cat}. In this case, the denoising process becomes:
\begin{equation}
\begin{split}
    &\mathrm{d}\boldx = \bigg[\bm{f}(\boldx, t) - g(t)^{2} \Big(\guidedcolor{\estscoreori(\boldx, t)} \ + \\
    &\quad \contrastcolor{\lambda_t \big( \estscoreguide(\boldx, t, \boldy^{+}) - \estscoreguide(\boldx, t, \boldy^{-}) \big)}\Big) \bigg]\mathrm{d}t + g(t)\mathrm{d}\bar{\boldw}.
\end{split}
\end{equation}

\paragraph{Continuous, Rig-like Control} The text-only input of text-to-image diffusion models does not allow continuous, rig-like control because it cannot be accurately described by the text.
To gain this control, we can use \ourmethod and change $\lambda_t$ continuously. For instance, when the prompt $\boldy$ is \textit{a photo of a waterfall}, the positive prompt $\boldy^{+}$ can be \textit{a beautiful photo of a waterfall}, and the negative prompt $\boldy^{-}$ can be\textit{a not beautiful photo of a waterfall}. In these cases, the denoising process becomes:\footnote{\label{footnote:generalist-cfg}Previous works \cite{Rombach2021ldm} have shown that text-to-image models need classifier-free guidance (CFG) \cite{ho2021classifierfree} to generate reasonable output. In this paper, $\guidedcolor{\tilde{\estscoreguide}}$ refers to the score function after CFG. See details in Appendices~\ref{app:guide-generalist} and \ref{subapp:baselines-guide-expert}.} 
\begin{equation}
\begin{split}
    &\mathrm{d}\boldx = \bigg[\bm{f}(\boldx, t) - g(t)^{2} \Big(\guidedcolor{\tilde{\estscoreguide}(\boldx, t, \boldy)} \ + \\
    &\quad \contrastcolor{\lambda_t \big( \estscoreguide(\boldx, t, \boldy^{+}) - \estscoreguide(\boldx, t, \boldy^{-}) \big)}\Big) \bigg]\mathrm{d}t + g(t)\mathrm{d}\bar{\boldw}.
\end{split}
\end{equation}
One can tune the value of $\lambda_t$ to gain continuous, rig-like control for text-to-image diffusion models. Specifically, when $\lambda_t$ is positive, the generation is guided towards the positive prompt; when $\lambda_t$ is negative, the generation is guided towards the opposite direction of the positive prompt. 

\begin{figure*}[!t]
\centering
  \includegraphics[width=\linewidth]{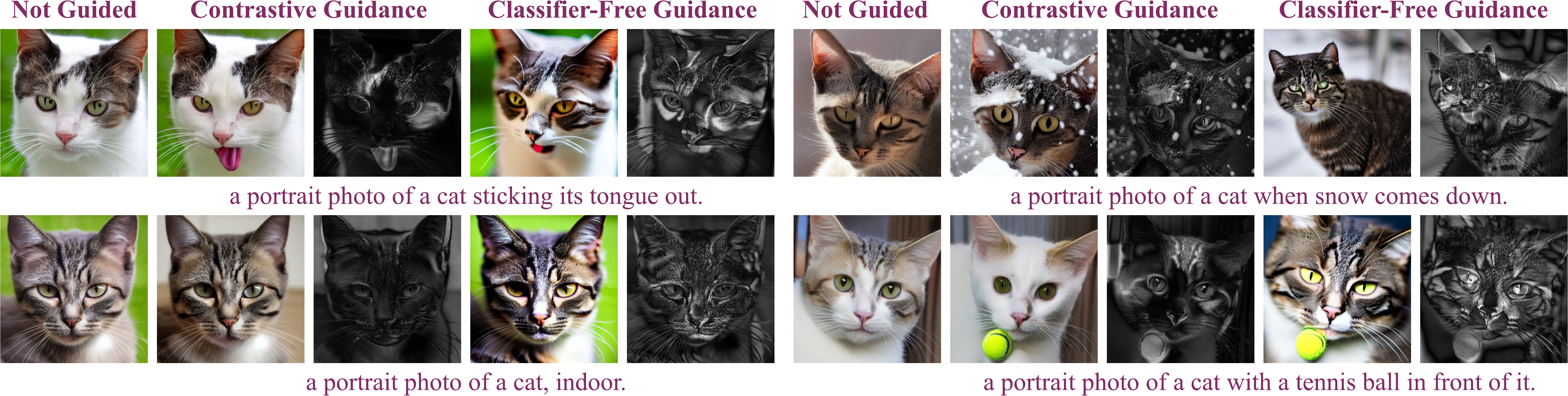}
\caption{\label{fig:visualization} \ourmethod improves disentanglement (e.g., attributes, background, foreground, and objects). Within each row, we fixed all random variables during the sampling process. The gray-scale images visualize the norm of the pixel-wise distance between the two images before and after the guidance.
}
\end{figure*}

\paragraph{Guiding Image Editors} Zero-shot image editing aims at modifying a real image $\boldx$ given a source text $\boldy$ and a target text $\hat{\boldy}$ \cite{unify}. For instance, suppose one wants to change the season of a real image, taken from an aerial view, from autumn to winter. The source text $\boldy$ can be \textit{an aerial view of autumn scene}, and the target text $\hat{\boldy}$ can be \textit{an aerial view of winter scene}. Since disentangling image factors is a key to meaningful editing, \ourmethod can be helpful to this task. In this paper, we focus on two editors: SDEdit \cite{meng2022sdedit} and \cyclediff \cite{unify}. SDEdit is based on the finding that, when noises are added and then removed up to a suitable noise level, the structure of the image can preserve. For zero-shot image editing, SDEdit first adds noise to the source image $\boldx$ to get a noisy image $\boldx_{t_{\text{SD}}}$. Then, conditioned on the target text $\hat{\boldy}$, it denoises $\boldx_{t_{\text{e}}}$ into a target image $\hat{\boldx}$ via
\begin{equation}
\label{eq:decoding-sde}
    \mathrm{d}\hat{\boldx} = \Big[\bm{f}(\hat{\boldx}, t) - g(t)^{2} \guidedcolor{\tilde{\estscoreguide}(\hat{\boldx}, t, \hat{\boldy})} \Big]\mathrm{d}t + g(t)\mathrm{d}\bar{\boldw}.
\end{equation}
\cyclediff leverages the observation that fixing all random variables in the diffusion process helps produce similar images. Similar to SDEdit, \cyclediff also adds noise to get $\boldx_{t_{\text{e}}}$ and denoises it with Eq.~(\ref{eq:decoding-sde}), but the difference is that the random variable $\mathrm{d}\bar{\boldw}$ in the denoising process in Eq.~(\ref{eq:decoding-sde}) is not randomly sampled. Instead, \cyclediff introduces an encoding process that infers the $\mathrm{d}\bar{\boldw}$ that are likely to help reconstruct the source image given the source text. In the denoising process, the encoded $\mathrm{d}\bar{\boldw}$ are applied to Eq.~(\ref{eq:decoding-sde}). This helps improve content preservation during editing. \ourmethod can strengthen the intended editing for both methods by modifying the denoising process in Eq.~(\ref{eq:decoding-sde}). Specifically, we set the positive prompt $\boldy^{+}$ as the target text $\hat{\boldy}$ and the baseline prompt $\boldy^{-}$ as the source text $\boldy$. In this case, the denoising process now becomes:
\begin{equation}
\label{eq:sde-decoding-contrastive}
\begin{split}
    \mathrm{d}\hat{\boldx} &= \Big[\bm{f}(\hat{\boldx}, t) - g(t)^{2} \Big(\guidedcolor{\tilde{\estscoreguide}(\hat{\boldx}, t, \hat{\boldy})} \ + \\
    &\quad\ \  \contrastcolor{\lambda_t \big(\estscoreguide(\hat{\boldx}, t, \hat{\boldy}) - \estscoreguide(\hat{\boldx}, t, \boldy)\big)} \Big) \Big]\mathrm{d}t + g(t)\mathrm{d}\bar{\boldw}.
\end{split}
\end{equation}

\begin{table}[!t]
    \centering
    \begin{adjustbox}{width=\linewidth}
    \begin{tabular}{@{}lccc@{}}
        \toprule
                                    & AFHQ Cat  & AFHQ Dog  & FFHQ \\
        \midrule
        Classifier-Free Guidance    & 0.341     & 0.287     & 0.118 \\
        \ourmethod (ours)           & \bf 0.271 & \bf 0.265 & \bf 0.104 \\
        \bottomrule
    \end{tabular}
    \end{adjustbox}\smallskip
\caption{\label{tab:diff} \ \ \ourmethod improves disentanglement. The metric is the average pixel-wise $L_2$ distance between the two images before and after the guidance. }
\end{table}

\begin{table*}[!th]
    \centering
    \begin{adjustbox}{width=0.95\linewidth}
    \begin{tabular}{@{}lcccccccc@{}}
        \toprule
        & \multicolumn{3}{c}{AFHQ Cat} & \multicolumn{3}{c}{AFHQ Dog} & \multicolumn{2}{c}{FFHQ} \\
        \cmidrule(lr){2-4} \cmidrule(lr){5-7} \cmidrule(l){8-9}
        Method & FID\down & KID $\times 10^3$\down & $\mathcal{S}_{\text{CLIP}}$\up & FID\down & KID $\times 10^3$\down & $\mathcal{S}_{\text{CLIP}}$\up & FID\down & $\mathcal{S}_{\text{CLIP}}$\up \\
        \midrule
        Text-to-Image Only                             & 48.5      & 35.2      & 0.327     & 83.4      & 51.4      & 0.328     & 69.8      & \bf 0.305 \\
        Finetuned Text-to-Image Only                   & 41.3      & 35.2      & 0.316     & 67.1      & 38.9      & 0.319     & 47.3      & 0.280 \\
        Domain Expert + Classifier-Free Guidance    & 43.9      & 39.5      & 0.311     & 78.0      & 51.6      & 0.315     & 50.0      & 0.279 \\
        Domain Expert + Negative Guidance           & 124.6     & 110.7     & 0.230     & 94.5      & 60.7      & 0.239     & 168.1     & 0.175 \\
        Domain Expert + \ourmethod                  & \bf 32.0  & \bf 18.8  & \bf 0.333 & \bf 52.8  & \bf 27.0  & \bf 0.329 & \bf 42.2  & 0.280 \\
        \bottomrule
    \end{tabular}
    \end{adjustbox}\smallskip
    \caption{\label{tab:domain-results} \ourmethod improve domain-specific text-to-image modeling.  }
\end{table*}

\section{Disentanglement Analysis}
\label{sec:visualization}

In this section, we show that \ourmethod provides more disentangled guidance than classifier-free guidance \cite{ho2021classifierfree} (\cfg; for this baseline, the positive prompt is used as input). To study disentanglement, we study the pixel-wise $L_2$ distance between pairs of images before the guidance (i.e., the domain expert) and after the guidance (i.e., \ourmethod or CFG). All random variables are fixed during sampling. Figure~\ref{fig:visualization} shows samples and visualizations of the $L_2$ distance. We can see regions of the tongue, the background, the snowflakes, and the tennis ball clearly when using \ourmethod, which are not observed when using \cfg. Table~\ref{tab:diff} also quantitatively verifies this observation. These results show that the guidance by \ourmethod is more disentangled than \cfg. 

\section{Applications}
\label{sec:experiments}

This section provides experiments of \ourmethod in several applications: (1) guiding domain experts for domain-specific text-to-image synthesis (Section~\ref{subsec:guide-expert}), (2) guiding text-to-image models to correct their own mistakes and to gain continuous, rig-like controls (Section~\ref{subsec:guide-generalist}), and (3) guiding image editors to strengthen the intended edit (Section~\ref{subsec:cycle-experiment}).

\subsection{Guiding Domain Experts}
\label{subsec:guide-expert}

\paragraph{Datasets:} We used three datasets: AFHQ Cat \cite{choi2020starganv2}, AFHQ Dog \cite{choi2020starganv2}, and FFHQ \cite{Karras2019ASG}. For each dataset, we created a set of 64 text pairs $(\boldy^{+}, \boldy^{-})$ (see Appendix~\ref{subsec:text-prompts-guide-expert}). The training sets of these datasets are used as (1) training data for domain experts and (2) references for FID and KID, which we will detail later. The validation sets are not used here. 

\paragraph{Pre-trained models: } We used the \texttt{SD-v1-5} Stable Diffusion model \cite{Rombach2021ldm} as the text-to-image diffusion model. Since Stable Diffusion is a latent diffusion model, we need the domain expert to share the encoder and decoder with it. To this effect, one can use the encoder of Stable Diffusion to encode images in each of the three datasets and then train a UNet on the encoded representations. To save computation, we finetune a copy of the text-to-image diffusion model into a domain expert:\footnote{\label{footnote:finetune} In practice, sometimes data owners may not want to share domain data for finetuning. Instead, they can still provide a domain expert to be guided. In these cases, the \textit{Finetuned Text-to-Image Only} baseline does not apply, while our \ourmethod still applies. }
we used a fixed text input: \textit{A portrait photo of [domain] *}, where \textit{[domain]} is the domain identifier, e.g., \textit{cat}. The UNet and the embedding of \textit{*} are finetuned, and the latter has a larger learning rate. The described finetuning method can be viewed as a combination of textual inversion \cite{Gal2022AnII} and DreamBooth \cite{Ruiz2022DreamBoothFT}.

\paragraph{Baselines: } (1) \textit{Text-to-Image Only} samples images from the text-to-image diffusion model, conditioned on $\boldy^{+}$, (2) \textit{Finetuned Text-to-Image Only}\footref{footnote:finetune} samples images from the finetuned text-to-image diffusion model, conditioned on $\boldy^{+}$, (3) \textit{Classifier-Free Guidance} \cite{ho2021classifierfree} leverages the positive prompt $\boldy^{+}$ to guide the domain expert, and (4) \textit{Expert + Negative Guidance} \cite{Liu2022CompositionalVG} uses the baseline prompt $\boldy^{-}$ to negatively guide the domain expert. 

\paragraph{Metrics: } For each text pair, we sample 16 samples from each method (1024 samples for each method in total). To evaluate the realism and domain specificity, we used Clean FID \cite{Parmar2021OnAR} to compute the Frechet Inception Distance (FID \cite{Heusel2017GANsTB}) and Kernel Inception Distance (KID \cite{bikowski2018demystifying}) of the 1024 samples using the training set as reference. On AFHQ, we evaluated with resolution $512 \times 512$, while on FFHQ, we evaluated with resolution $256 \times 256$, both with pre-computed statistics by \cite{Parmar2021OnAR}. Since statistics are not available for KID on FFHQ, we omitted this term. To evaluate the closeness to the text, we reported the CLIP score, i.e., the cosine similarity between the normalized CLIP embeddings of ${\boldx}$ and $\boldy^{+}$:
\begin{equation}
    \mathcal{S}_{\text{CLIP}}({\boldx}, \boldy^{+}) = \cos\big\langle\text{CLIP}_{\text{img}}({\boldx}), \text{CLIP}_{\text{text}}(\boldy^{+})\big\rangle.
\end{equation}

Table~\ref{tab:domain-results} demonstrates that \ourmethod achieves the best realism and domain specificity (i.e., FID and KID); also, images from our method have higher or comparable closeness to the positive text (i.e., $\mathcal{S}_{\text{CLIP}}$) compared to \textit{Text-to-Image Only}, which ignores domain specificity. 

\begin{figure*}[!th]
\centering
  \includegraphics[width=\linewidth]{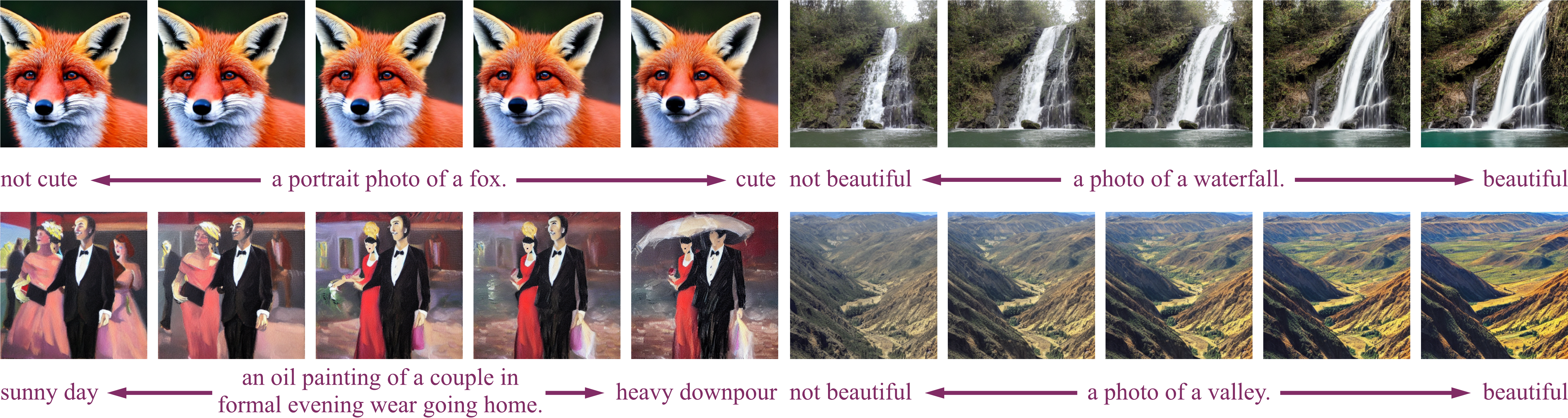}
\caption{\label{fig:rig-experiment} \ourmethod for continuous, rig-like controls, which cannot be described accurately by language. Results show that \ourmethod can guide text-to-image diffusion models to generate images with continuously changed variations. The positive and baseline prompts are provided in Table~\ref{tab:rig-prompt-examples}. Images are resized to $256 \times 256$, and please zoom in to see the details. }
\end{figure*}

\begin{figure}[!th]
\centering
  \includegraphics[width=\linewidth]{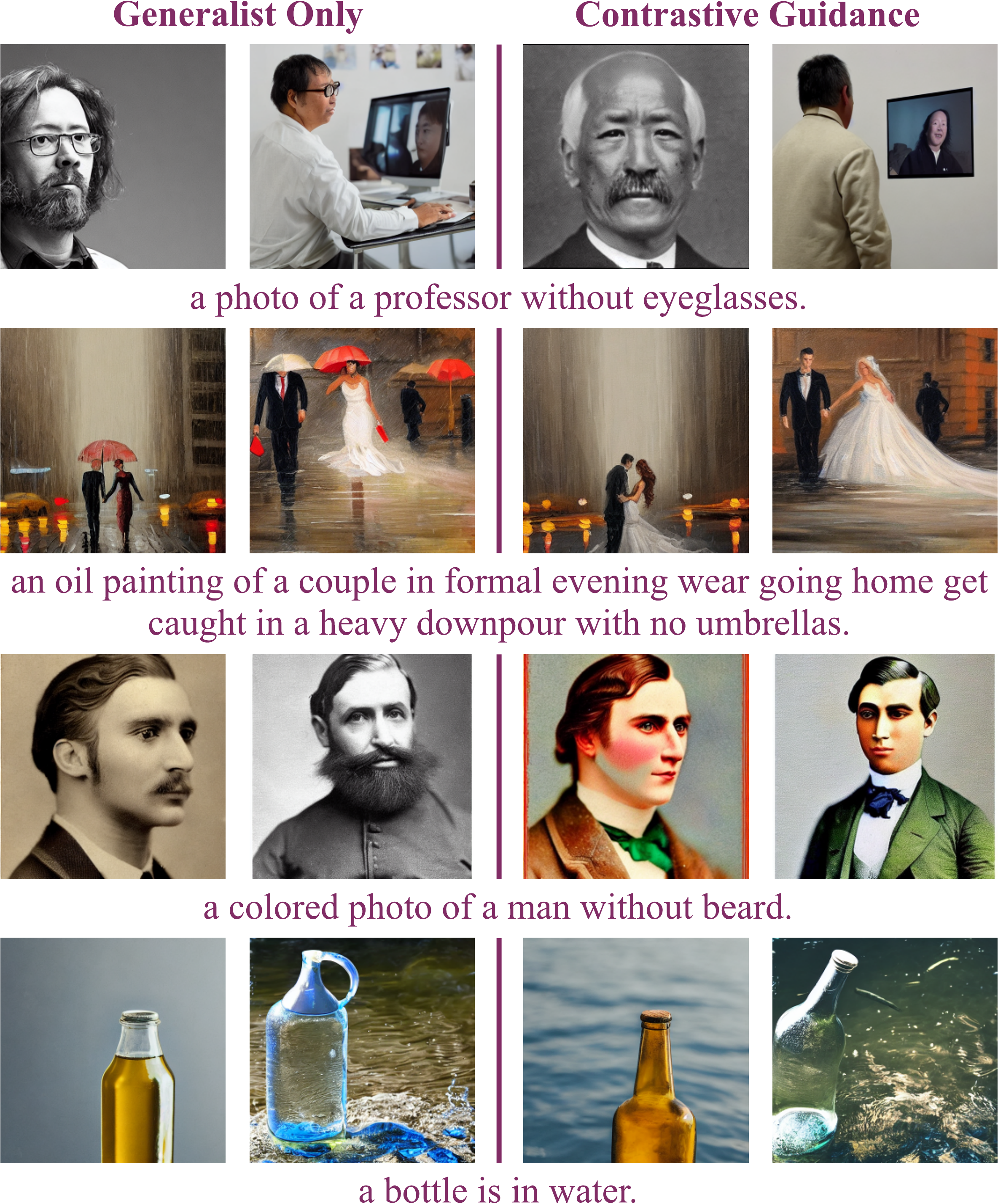}
\caption{\label{fig:hard-experiment} \ourmethod mitigate contradiction. We focused on failure cases of the text-to-image diffusion model such as understanding negation (\textit{without eyeglasses}, \textit{no umbrellas}, \textit{without beard}), relation (\textit{in water}), and one unexpected contradiction (\textit{colored photo}). The prompts used in this experiment are provided in Table~\ref{tab:hard-prompt-examples}. }
\end{figure}

\subsection{Guiding Text-to-Image Models}
\label{subsec:guide-generalist}

In this section, we use \ourmethod to help a text-to-image diffusion model (\texttt{SD-v1-5}, following Section~\ref{subsec:guide-expert}) to guide itself for negation, relations, and continuous, rig-like controls. The prompts we used are provided in Appendix~\ref{app:guide-generalist}. Since vision-language models like CLIP are not reliable for verifying text-image consistency in challenging cases \cite{Thrush2022WinogroundPV}, we provide image samples in this section. 

Figure~\ref{fig:hard-experiment} shows that \ourmethod improves the consistency between the synthesized images and some text prompts with negation, relation, and unlikely scenarios. For instance, the text-to-image diffusion model tends to synthesize contradictory images when the text contains \textit{without eyeglasses}, \textit{a heavy downpour with no umbrellas} \cite{Marcus2022AVP}, \textit{colored photo}, \textit{without beard}, and \textit{a bottle in water} \cite{Thrush2022WinogroundPV}. We leveraged \ourmethod to mitigate such contradiction by specifying baseline prompts such as \textit{with eyeglasses} and \textit{water in a bottle}. 

Figure~\ref{fig:rig-experiment} shows that \ourmethod enables continuous, rig-like controls ($\lambda_t \in \{-8, -4, 0, 4, 8\}$) such as the cuteness of the fox, the polarity of the weather (e.g., sunny \textit{vs.} rainy), and the aesthetics of the scene. The positive and baseline prompts used in this experiment are provided in Table~\ref{tab:rig-prompt-examples}.

\begin{figure*}[!th]
\centering
  \includegraphics[width=\linewidth]{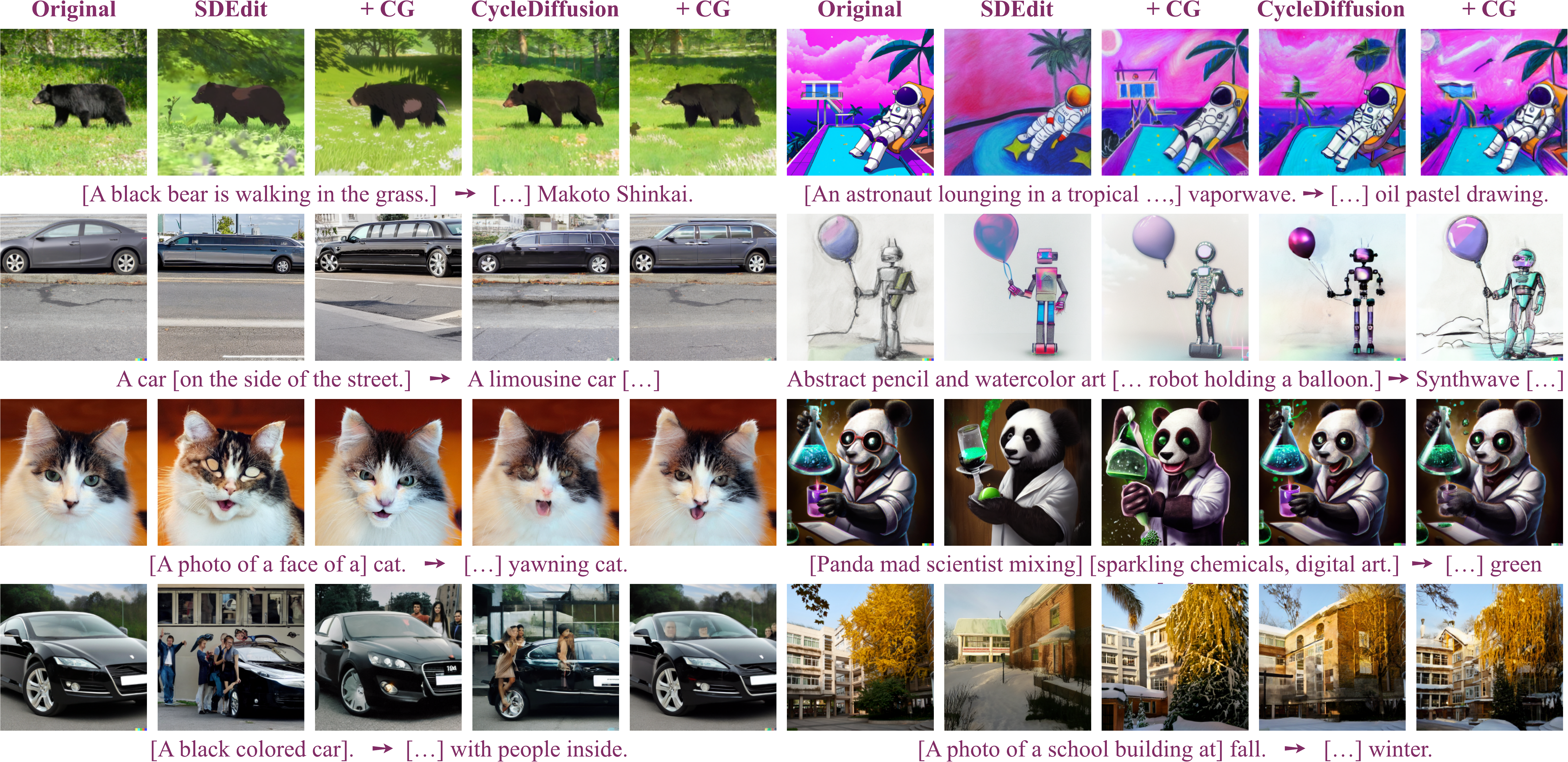}
\caption{\label{fig:cycle-experiment} \ourmethod can improve zero-shot image editing methods such as SDEdit \cite{meng2022sdedit} and \cyclediff \cite{unify}. We did not use a fixed combination of hyperparameters because every input can have its best combination of hyperparameters and even random seed. Following \cite{unify}, for each input, we ran 15 random trials for each combination of hyperparameters and report the one with the highest $\mathcal{S}_{\text{D-CLIP}}$. All images are resized to $256 \times 256$, and please zoom in to see the details.
}
\end{figure*}

\subsection{Guiding Image Editors}
\label{subsec:cycle-experiment}

This section demonstrates that \ourmethod can improve zero-shot image editors such as SDEdit and \cyclediff. The input is a triplet $(\boldx, \boldy, \hat{\boldy})$, where $\boldx$ is the source image, $\boldy$ is the source text, and $\hat{\boldy}$ is the target text; the output is a target image $\hat{\boldx}$ to be generated. As detailed in Section~\ref{subsec:application-method}, we apply \ourmethod to the decoding of SDEdit and \cyclediff, by setting the positive prompt $\boldy^{+}$ as the target text $\hat{\boldy}$ and the baseline prompt $\boldy^{-}$ as the source text $\boldy$. Please find more details in Appendix~\ref{app:cycle}. 

\paragraph{Datasets and pre-trained diffusion models: } 
We used the set of 150 triplets $({\boldx}, \boldy, \hat{\boldy})$ in \cite{unify}, covering editing of image styles, attributes, and objects. We used the \texttt{SD-v1-5} Stable Diffusion as the text-to-image diffusion model. 

\paragraph{Metrics: } We reported (1) \psnr~and SSIM, which evaluate the faithfulness of the edited images to the source images (i.e., minimality of editing); (2) CLIP score, which evaluates the authenticity of the edited images to the target text: 
\begin{equation}
    \mathcal{S}_{\text{CLIP}}({\hat{\boldx}}, \hat{\boldy}) =  \cos\big\langle\text{CLIP}_{\text{img}}({\hat{\boldx}}), \text{CLIP}_{\text{text}}(\hat{\boldy})\big\rangle;
\end{equation}
(3) directional CLIP score used by \cite{unify}:
\begin{equation}
\label{eq:directional-clip-score}
\begin{split}
    & \mathcal{S}_{\text{D-CLIP}}({\hat{\boldx}}, {\boldx}, \boldy, \hat{\boldy}) \ =  \\ 
    & \quad\quad\quad\ \ \cos\Big\langle\text{CLIP}_{\text{img}}({\hat{\boldx}}) - \text{CLIP}_{\text{img}}({\boldx}), \\
    & \quad\quad\quad\quad\quad\quad\quad\quad\quad \text{CLIP}_{\text{text}}(\hat{\boldy}) - \text{CLIP}_{\text{text}}(\boldy) \Big\rangle.
\end{split}
\end{equation}

Seen in Table~\ref{tab:dalle-results} are results for zero-shot image-to-image translation. For each method (SDEdit and \cyclediff), using \ourmethod results in higher performances in terms of all metrics. Also, the use of \ourmethod allows us to use a less number of hyperparameter enumerations and random trials for all methods as is done in \cite{unify} (see Appendix~\ref{app:cycle}), making the three methods nearly $4\times$ more efficient while achieving higher performances. 

\begin{table}[!t]
    \centering
    \begin{adjustbox}{width=0.98\linewidth}
    \begin{tabular}{@{}lcccc@{}}
        \toprule
        & $\mathcal{S}_{\text{CLIP}}$\up & $\mathcal{S}_{\text{D-CLIP}}$\up & \psnr\up & SSIM\up \\
        \midrule
        SDEdit                      & \bf 0.344 & 0.258     & 15.93     & 0.512 \\
        SDEdit + \textit{CG}        & 0.336     & \bf 0.275 & \bf 17.37 & \bf 0.553 \\
        \midrule
        \cyclediff                  & \bf 0.334 & 0.272     & 21.92     & 0.731 \\
        \cyclediff + \textit{CG}    & 0.330     & \bf 0.278 & \bf 22.65 & \bf 0.742 \\
        \bottomrule
    \end{tabular}
    \end{adjustbox}\smallskip
\caption{\label{tab:dalle-results} \ourmethod (denoted as + \textit{CG}) can improve SDEdit and \cyclediff for zero-shot image editing. We did not use a fixed combination of hyperparameters, and neither did we plot the trade-off curve. The reason is that every input can have its best combination of hyperparameters and even random seed. Instead, following \cite{unify}, for each input, we ran 15 random trials for each combination of hyperparameters and report the one with the highest $\mathcal{S}_{\text{D-CLIP}}$. Please find more details in Appendix~\ref{app:cycle}. }
\end{table}

\section{Conclusions and Discussion}
\label{sec:conclusion}

This paper advocates the idea of using text-to-image diffusion models as disentangled guidance for image synthesis. The key idea is to disentangle image factors using two prompts that have minimal differences. We formalize this intuition by defining a generative classifier with the text-to-image model and the two prompts, whose score function is proportional to a simple subtraction term. We have shown the efficacy of \ourmethod in several applications: (1) guiding domain experts for domain-specific but generalizable text-to-image synthesis, (2) guiding text-to-image models themselves to gain continuous, rig-like controls, and (3) guiding image editors to improve the intended edit. Our experiments show that \ourmethod is both qualitatively and quantitatively better than the baselines. 

This paper also leaves several questions for further exploration. For efficiency, we assume that the temperature $\gamma_t$ is adaptive such that $\lambda_t$ does not depend on $\boldx$; although its effectiveness is validated by our experiments, it should be considered with reservation because the validity may not hold on other domains. Moreover, one may study how the choice of the prompt pairs $(\boldy^{+}, \boldy^{-})$ affects the performance. Please see our discussion on errors and biases in Appendix~\ref{subapp:error-bias-analysis}.




{
    \small
    \bibliographystyle{ieeenat_fullname}
    \bibliography{main}
}

\clearpage

\appendix 

{\onecolumn 
\section{Mathematical Details}
\label{app:math-details}

\subsection{Derivation of Score Function of Generative Classifier}
\label{subapp:derivation-score}

Here, we derive $\nabla_{\boldx} \log p_{t}(\boldx|c)$ of the generative classifier $p(\boldx|c)$ defined by \ourmethod. Recall that, given a positive prompt $\boldy^{+}$ and a baseline prompt $\boldy^{-}$, in Section~\ref{subsec:method-diffusion} we have rewritten this generative classifier as 
\begin{align}
    &p_{t}(\boldx|c) \propto p_{t}(\boldx) p_{t}(c|\boldx) = p_{t}(\boldx) \frac{p(\boldy^{+})p_{t}(\boldx|\boldy^{+})^{\gamma_t}}{p(\boldy^{+})p_{t}(\boldx|\boldy^{+})^{\gamma_t} + p(\boldy^{-})p_{t}(\boldx|\boldy^{-})^{\gamma_t}},
\end{align}
where the temperature $\gamma_t$ determines how sharp the generative classifier is. Given enough space, here we do not simplify notations as done in Section~\ref{subsec:method-diffusion}. Using the \textit{log-derivative trick} is used once in Eq.~(\ref{eq:trick}), we derive the score function as
\begin{align}
\begin{split}
    &\quad \ \nabla_{\boldx} \log p_{t}(\boldx|c) \\
    &= \nabla_{\boldx} \log \Big( p_{t}(\boldx) \frac{p(\boldy^{+})p_{t}(\boldx | \boldy^{+})^{\gamma_t}}{p(\boldy^{+})p_{t}(\boldx | \boldy^{+})^{\gamma_t} + p(\boldy^{-})p_{t}(\boldx | \boldy^{-})^{\gamma_t}} \Big)
\end{split}\\
    &=\nabla_{\boldx} \log p_{t}(\boldx) + 0 + \nabla_{\boldx} \log p_{t}(\boldx | \boldy^{+})^{\gamma_t} - \nabla_{\boldx} \log \Big( p(\boldy^{+})p_{t}(\boldx | \boldy^{+})^{\gamma_t} + p(\boldy^{-})p_{t}(\boldx | \boldy^{-})^{\gamma_t} \Big) \\
    &= \nabla_{\boldx} \log p_{t}(\boldx) + \nabla_{\boldx} \log p_{t}(\boldx | \boldy^{+})^{\gamma_t} - \frac{\nabla_{\boldx} \Big( p(\boldy^{+})p_{t}(\boldx | \boldy^{+})^{\gamma_t} + p(\boldy^{-})p_{t}(\boldx | \boldy^{-})^{\gamma_t} \Big)}{p(\boldy^{+})p_{t}(\boldx | \boldy^{+})^{\gamma_t} + p(\boldy^{-})p_{t}(\boldx | \boldy^{-})^{\gamma_t}}  \\
    \label{eq:trick}
\begin{split}
    &=\nabla_{\boldx} \log p_{t}(\boldx) + \nabla_{\boldx} \log p_{t}(\boldx | \boldy^{+})^{\gamma_t} \\
    &\quad\quad\quad\quad\quad\quad\quad\quad\quad - \frac{ p(\boldy^{+})p_{t}(\boldx | \boldy^{+})^{\gamma_t} \nabla_{\boldx} \log p_{t}(\boldx | \boldy^{+})^{\gamma_t} + p(\boldy^{-}) p_{t}(\boldx | \boldy^{-})^{\gamma_t} \nabla_{\boldx} \log p_{t}(\boldx | \boldy^{-})^{\gamma_t} }{p(\boldy^{+})p_{t}(\boldx | \boldy^{+})^{\gamma_t} + p(\boldy^{-})p_{t}(\boldx | \boldy^{-})^{\gamma_t}} 
\end{split}\\
    &=\nabla_{\boldx} \log p_{t}(\boldx) + \frac{p(\boldy^{-})p_{t}(\boldx | \boldy^{-})^{\gamma_t}}{p(\boldy^{+})p_{t}(\boldx | \boldy^{+})^{\gamma_t} + p(\boldy^{-})p_{t}(\boldx | \boldy^{-})^{\gamma_t}} \Big(\nabla_{\boldx} \log p_{t}(\boldx | \boldy^{+})^{\gamma_t} - \nabla_{\boldx} \log p_{t}(\boldx | \boldy^{-})^{\gamma_t} \Big) \\
    \label{eq:contain-lambda-appendix}
    &=\nabla_{\boldx} \log p_{t}(\boldx) + \underbrace{\frac{\gamma_t \cdot p(\boldy^{-})p_{t}(\boldx | \boldy^{-})^{\gamma_t}}{p(\boldy^{+})p_{t}(\boldx | \boldy^{+})^{\gamma_t} + p(\boldy^{-})p_{t}(\boldx | \boldy^{-})^{\gamma_t}}}_{\lambda_t} \Big(\nabla_{\boldx} \log p_{t}(\boldx | \boldy^{+}) - \nabla_{\boldx} \log p_{t}(\boldx | \boldy^{-}) \Big) \\
    &=\nabla_{\boldx} \log p_{t}(\boldx) + \lambda_t \Big(\nabla_{\boldx} \log p_{t}(\boldx | \boldy^{+}) - \nabla_{\boldx} \log p_{t}(\boldx | \boldy^{-}) \Big) \\
    \label{eq:nn-approximation-appendix}
    &\approx \guidedcolor{\estscoreoriabstract(\boldx, t)} + \contrastcolor{\lambda_t \Big( \estscoreguide(\boldx, t, \boldy^{+}) - \estscoreguide(\boldx, t, \boldy^{-}) \Big)}. 
\end{align}

\subsection{Derivation of $\lambda_t$}
\label{subapp:estimate-lambda}

In Section~\ref{subsec:method-diffusion} and Appendix~\ref{subapp:derivation-score}, we have derived the score function $\nabla_{\boldx} \log p_{t}(\boldx|c)$ of the generative classifier $p(\boldx|c)$ defined by \ourmethod. However, a remaining question is how to estimate the coefficient $\lambda_t$ in Eq.~(\ref{eq:contain-lambda}) and Eq.~(\ref{eq:contain-lambda-appendix}). In the expression of $\lambda_t$, $\gamma_t$ is a hyperparameter, and $p(\boldy^{+})$ and $p(\boldy^{-})$ are priors of text prompts; therefore, to estimate $\lambda_t$, we only need to estimate $p_{t}(\boldx | \boldy^{+})$ and $p_{t}(\boldx | \boldy^{-})$. In the following, we show how to estimate $p_{t}(\boldx | \boldy^{+})$, and the estimate of $p_{t}(\boldx | \boldy^{-})$ can be derived similarly by replacing $\boldy^{+}$ with $\boldy^{-}$. 

Recall that $p_{t}(\boldx | \boldy^{+})$ is the marginal distribution of $\boldx$ of the following SDE, with initial state distribution $p_T(\boldx)$:
\begin{equation}
\label{eq:derive-lambda-sde}
    \mathrm{d}\boldx = \Big[\bm{f}(\boldx, t) - g(t)^{2} \nabla_{\boldx} \log p_{t}(\boldx | \boldy^{+}) \Big]\mathrm{d}t + g(t)\mathrm{d}\bar{\boldw}.
\end{equation}
It has been shown that each SDE has a corresponding \textit{probability flow ODE} with the same marginal density at each time $t$ \cite{Maoutsa2020InteractingPS,0011SKKEP21}. Specifically, probability flow ODE of the SDE in Eq.~(\ref{eq:derive-lambda-sde}) is
\begin{equation}
\label{eq:derive-lambda-ode}
    \mathrm{d}\boldx = \Big[\bm{f}(\boldx, t) - \frac{1}{2} g(t)^{2} \nabla_{\boldx} \log p_{t}(\boldx | \boldy^{+}) \Big]\mathrm{d}t.
\end{equation}
Since the SDE and the probability flow ODE have the same marginal distribution, we can estimate $p_{t}(\boldx | \boldy^{+})$ by estimating the marginal $p_{t}^{\text{ODE}}(\boldx | \boldy^{+})$ of the probability flow ODE in Eq.~(\ref{eq:derive-lambda-ode}). Since $p_{T}^{\text{ODE}}(\boldx | \boldy^{+})$ is the isometric Gaussian distribution $\mathcal{N}(\boldx; \bm{0}, \bm{I})$, the marginal $p_{t}^{\text{ODE}}(\boldx | \boldy^{+})$ is tractable by integrating the following ODE \cite{Chen2018NeuralOD,Grathwohl2019FFJORDFC} from $t$ to $T$.
\begin{equation}
\label{eq:derive-lambda-density-ode}
    \mathrm{d}\begin{bmatrix}
        \boldx \\ 
        \log p_{t}^{\text{ODE}}(\boldx | \boldy^{+})
    \end{bmatrix}
    = \begin{bmatrix}
        \displaystyle \bm{f}(\boldx, t) - \frac{1}{2} g(t)^{2} \nabla_{\boldx} \log p_{t}(\boldx | \boldy^{+}) \\
        \displaystyle -\mathrm{tr}\Big(\frac{\mathrm{d}}{\mathrm{d}\boldx}  \big(\bm{f}(\boldx, t) - \frac{1}{2} g(t)^{2} \nabla_{\boldx} \log p_{t}(\boldx | \boldy^{+})\big) \Big)
    \end{bmatrix}\mathrm{d}t.
\end{equation}
The generalist text-to-image diffusion model approximates $\nabla_{\boldx} \log p_{t}(\boldx | \boldy^{+})$ with $\estscoreguide(\boldx, t, \boldy^{+})$, and Eq.~(\ref{eq:derive-lambda-density-ode}) becomes 
\begin{equation}
\label{eq:derive-lambda-density-ode-approx}
    \mathrm{d}\begin{bmatrix}
        \boldx \\ 
        \log p_{t}^{\text{ODE}}(\boldx | \boldy^{+})
    \end{bmatrix}
    = \begin{bmatrix}
        \displaystyle \bm{f}(\boldx, t) - \frac{1}{2} g(t)^{2} \estscoreguide(\boldx, t, \boldy^{+}) \\
        \displaystyle -\mathrm{tr}\Big(\frac{\mathrm{d}}{\mathrm{d}\boldx}  \big(\bm{f}(\boldx, t) - \frac{1}{2} g(t)^{2} \estscoreguide(\boldx, t, \boldy^{+})\big) \Big)
    \end{bmatrix}\mathrm{d}t.
\end{equation}
Although the above estimation is tractable, we need to run it for all $t$'s, resulting in $\mathcal{O}(N^2)$ complexity where $N$ is the number of discretization steps. 
Therefore, although $\lambda_t$ can be estimated as we show above, this paper tries to avoid such estimation. To this effect, we assume that the temperature $\gamma_t$ is adaptive such that $\lambda_t$ does not depend on $\boldx$, giving us $\mathcal{O}(N)$ complexity.

\section{Guiding Domain Experts (Section~\ref{subsec:guide-expert}): Experimental Details}
\label{app:guide-expert}

\subsection{Task Input} 
\label{subsec:text-prompts-guide-expert}

\ourmethod allows generalists to guide domain experts for domain-specific text-to-image synthesis. The input to this task is a pair of positive and baseline prompts $(\boldy^{+}, \boldy^{-})$ (note that some baselines can only take one of them as input). 
Tables~\ref{tab:cat-prompt-examples}-\ref{tab:ffhq-prompt-examples} provide several examples for $(\boldy^{+}, \boldy^{-})$. \anonymoustext{The full lists of samples will be provided in our code release.}\acceptedtext{The full lists of samples are provided in our code release.} 

\subsection{Baseline Details}
\label{subapp:baselines-guide-expert}

\noindent\textbf{Generalist Only: } Following the notations in the main text, \textit{Generalist Only} means sampling from the reverse-time SDE:
\begin{equation}
\begin{split}
    &\mathrm{d}\boldx = \bigg[\bm{f}(\boldx, t) - g(t)^{2} \guidedcolor{\tilde{\estscoreguide}(\boldx, t, \boldy^{+})} \bigg]\mathrm{d}t + g(t)\mathrm{d}\bar{\boldw},
\end{split}
\end{equation}
where $\guidedcolor{\tilde{\estscoreguide}(\boldx, t, \boldy^{+})}$ means the score function after classifier-free guidance \cite{ho2021classifierfree}. It means that \textit{Generalist Only} samples from: 
\begin{equation}
\begin{split}
    &\mathrm{d}\boldx = \bigg[\bm{f}(\boldx, t) - g(t)^{2} \Big(\tau_t\estscoreguide(\boldx, t, \boldy^{+}) - (\tau_t - 1)\estscoreguide(\boldx, t, \varnothing) \Big) \bigg]\mathrm{d}t + g(t)\mathrm{d}\bar{\boldw},
\end{split}
\end{equation}
where $\varnothing$ is the null token that stands for unconditional sampling, and $\tau_t$ is a hyperparameter indicating the strength of guidance. We encourage readers to read \cite{ho2021classifierfree} for more details of classifier-free guidance. 

\noindent\textbf{Positive Guidance: } \textit{Positive Guidance} is essentially classifier-free guidance for the domain expert, using the positive prompt. Specifically, this baseline samples from the following reverse-time SDE:
\begin{equation}
\begin{split}
    &\mathrm{d}\boldx = \bigg[\bm{f}(\boldx, t) - g(t)^{2} \Big(\guidedcolor{\estscoreori(\boldx, t)} + \tau_t \big( \estscoreguide(\boldx, t, \boldy^{+}) - \estscoreguide(\boldx, t, \varnothing) \big)\Big) \bigg]\mathrm{d}t + g(t)\mathrm{d}\bar{\boldw},
\end{split}
\end{equation}
where $\varnothing$ is the null token that stands for unconditional sampling, and $\tau_t$ is a hyperparameter indicating the strength of guidance.

\noindent\textbf{Negative Guidance: } \textit{Negative Guidance} is essentially the negation operation \cite{Liu2022CompositionalVG} applied to the domain expert. Specifically, this baseline samples from the following reverse-time SDE:
\begin{equation}
\begin{split}
    &\mathrm{d}\boldx = \bigg[\bm{f}(\boldx, t) - g(t)^{2} \Big(\guidedcolor{\estscoreori(\boldx, t)} + \tau_t \big( \estscoreguide(\boldx, t, \varnothing) - \estscoreguide(\boldx, t, \boldy^{-}) \big)\Big) \bigg]\mathrm{d}t + g(t)\mathrm{d}\bar{\boldw},
\end{split}
\end{equation}
where $\varnothing$ is the null token that stands for unconditional sampling, and $\tau_t$ is a hyperparameter indicating the strength of guidance.

\noindent\textbf{StyleCLIP: } We used the latent optimization variant of StyleCLIP \cite{Patashnik2021StyleCLIPTM} as our baseline. Given the positive prompt $\boldy^{+}$, StyleCLIP begins by randomly sampling a style vector from a StyleGAN \cite{Karras2019ASG,Karras2020AnalyzingAI,karras2021aliasfree}, followed by iteratively optimizing this style vector to minimize the CLIP embedding distance between the generated image and the positive prompt $\boldy^{+}$. We used the implementation in the Google Colab Notebook provided by the authors of the original paper. 

\noindent\textbf{StyleGAN-NADA: } Similar to StyleCLIP, StyleGAN-NADA \cite{Gal2021StyleGANNADACD} also leverages the CLIP model to help generalize to text conditions, while StyleGAN-NADA finetunes the StyleGAN generator for each positive prompt $\boldy^{+}$ and keeps the style vector fixed. We used the implementation in the Google Colab Notebook provided by the authors of the original paper. 

\begin{table}[!ht]
    \centering
    \begin{tabular}{@{}ll@{}}
        \toprule
        Positive prompt $\boldy^{+}$ & Baseline prompt $\boldy^{-}$ \\
        \midrule
        a portrait photo of a cat with eyeglasses. & a portrait photo of a cat. \\
        a portrait photo of a cat, indoor. & a portrait photo of a cat, outdoor. \\
        a portrait photo of a yellow cat.         & a portrait photo of a cat. \\
        a portrait photo of a hairy cat. & a portrait photo of a hairless cat. \\
        a portrait photo of a cat with open mouth. & a portrait photo of a cat with closed mouth. \\
        a portrait photo of a cat at night. & a portrait photo of a cat in the daytime. \\
        a portrait photo of a cat with fireworks behind. & a portrait photo of a cat. \\
        a portrait photo of a cat with a tennis ball in front of it. & a portrait photo of a cat. \\
        a portrait photo of a cat when rain comes down. & a portrait photo of a cat. \\
        a portrait photo of a British Shorthair cat. & a portrait photo of a cat. \\
        a portrait photo of a cat with snow. & a portrait photo of a cat without snow. \\
        a portrait photo of a cat when snow comes down. & a portrait photo of a cat. \\
        a portrait photo of a cat underwater. & a portrait photo of a cat. \\
        \bottomrule
    \end{tabular}\smallskip
\caption{\label{tab:cat-prompt-examples} Examples of text prompts for AFHQ Cat (Section~\ref{subsec:guide-expert}). See image samples from our method in Figures~\ref{fig:cat1}-\ref{fig:cat3}. }
\end{table}

\begin{table}[!ht]
    \centering
    \begin{tabular}{@{}ll@{}}
        \toprule
        Positive prompt $\boldy^{+}$ & Baseline prompt $\boldy^{-}$ \\
        \midrule
        a portrait photo of a dog with eyeglasses. & a portrait photo of a dog. \\
        a portrait photo of a dog, indoor. & a portrait photo of a dog, outdoor. \\
        a portrait photo of a yellow dog.         & a portrait photo of a dog. \\
        a portrait photo of a hairy dog. & a portrait photo of a hairless dog. \\
        a portrait photo of a dog with open mouth. & a portrait photo of a dog with closed mouth. \\
        a portrait photo of a dog at night. & a portrait photo of a dog in the daytime. \\
        a portrait photo of a dog with fireworks behind. & a portrait photo of a dog. \\
        a portrait photo of a dog with a tennis ball in front of it. & a portrait photo of a dog. \\
        a portrait photo of a dog when rain comes down. & a portrait photo of a dog. \\
        a portrait photo of a German Shepherd dog. & a portrait photo of a dog. \\
        a portrait photo of a dog with snow. & a portrait photo of a dog without snow. \\
        a portrait photo of a snow when snow comes down. & a portrait photo of a dog. \\
        a portrait photo of a dog underwater. & a portrait photo of a dog. \\
        \bottomrule
    \end{tabular}\smallskip
\caption{\label{tab:dog-prompt-examples} Examples of text prompts for AFHQ Dog (Section~\ref{subsec:guide-expert}). }
\end{table}

\begin{table}[!ht]
    \centering
    \begin{tabular}{@{}ll@{}}
        \toprule
        Positive prompt $\boldy^{+}$ & Baseline prompt $\boldy^{-}$ \\
        \midrule
        a portrait photo of a person with eyeglasses. & a portrait photo of a person. \\
        a portrait photo of a person, indoor. & a portrait photo of a person, outdoor. \\
        a portrait photo of a black woman.         & a portrait photo of a person. \\
        a portrait photo of a hairy person. & a portrait photo of a hairless person. \\
        a portrait photo of a person with open mouth. & a portrait photo of a person with closed mouth. \\
        a portrait photo of a person at night. & a portrait photo of a person in the daytime. \\
        a portrait photo of a person with fireworks behind. & a portrait photo of a person. \\
        a portrait photo of a person with a tennis ball in front of it. & a portrait photo of a person. \\
        a portrait photo of a person when rain comes down. & a portrait photo of a person. \\
        a portrait photo of a person with a headphone. & a portrait photo of a person without a headphone. \\
        a portrait photo of a person with snow. & a portrait photo of a person without snow. \\
        a portrait photo of a person when snow comes down. & a portrait photo of a person. \\
        a portrait photo of a person underwater. & a portrait photo of a person. \\
        \bottomrule
    \end{tabular}\smallskip
\caption{\label{tab:ffhq-prompt-examples} Examples of text prompts for FFHQ (Section~\ref{subsec:guide-expert}). }
\end{table}

\begin{figure*}[!ht]
\centering
  \includegraphics[width=0.85\linewidth]{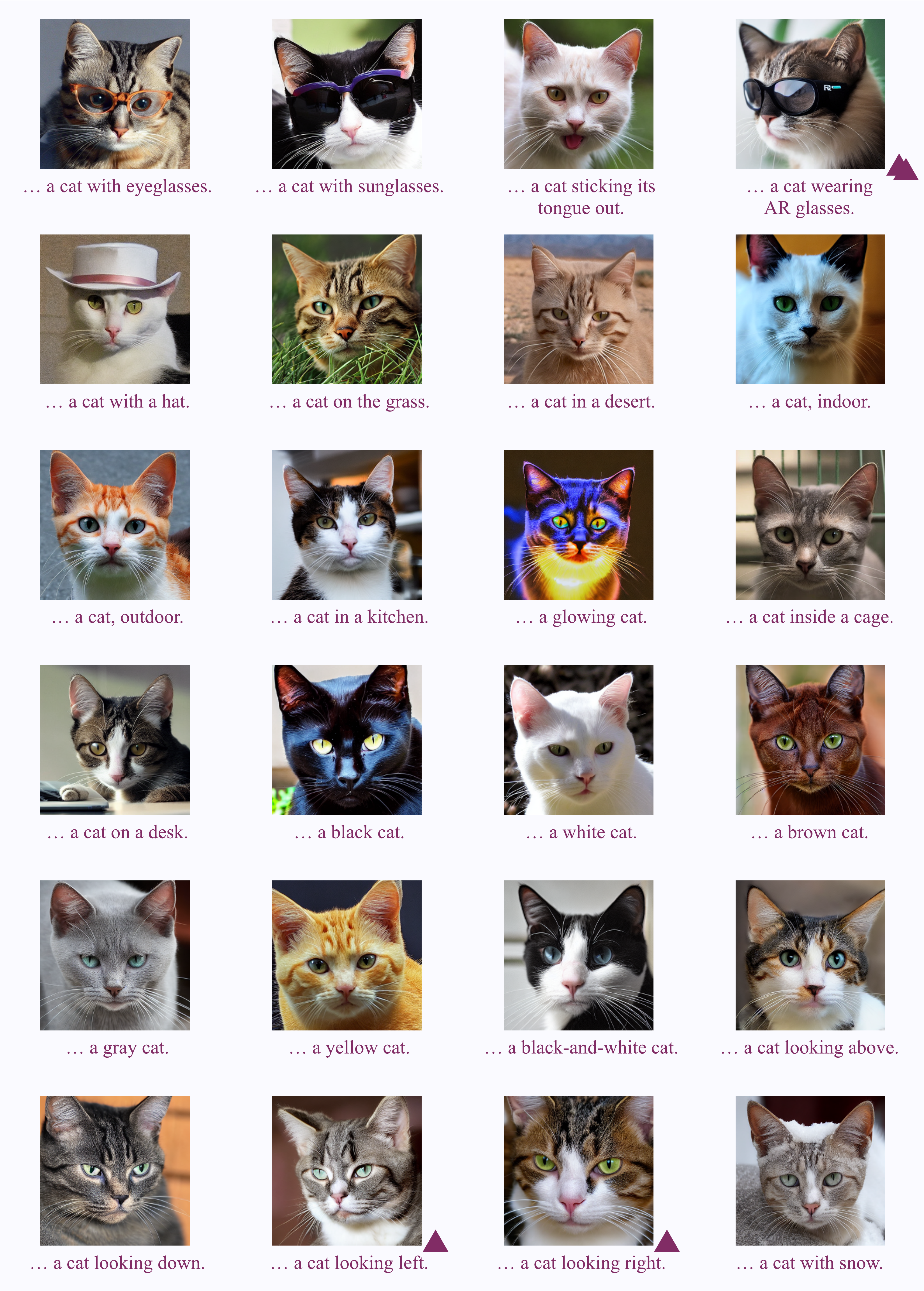}
\caption{\label{fig:cat1} Samples for text-to-image synthesis in the AFHQ Cat domain, using our \ourmethod. A single triangle beside an image stands for \textit{it fails sometimes}, while double triangles beside an image stand for \textit{it usually fails}. The positive text $\boldy^{+}$ is displayed below each image, in which $\ldots$ stands for ``a portrait photo of''. 
}
\end{figure*}

\begin{figure*}[!ht]
\centering
  \includegraphics[width=0.85\linewidth]{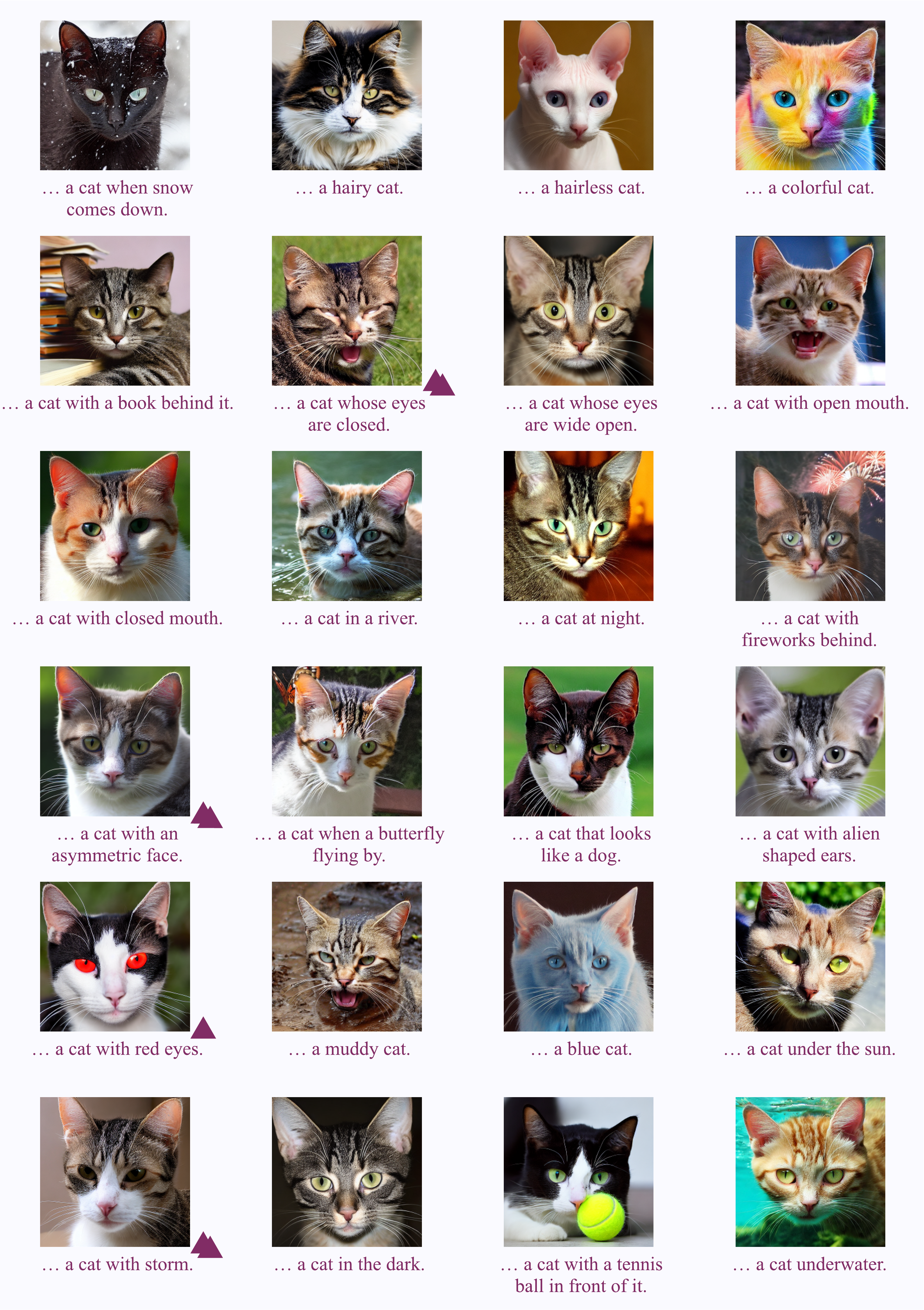}
\caption{\label{fig:cat2} Samples for text-to-image synthesis in the AFHQ Cat domain, using our \ourmethod. Notations follow Figure~\ref{fig:cat1}.
}
\end{figure*}

\begin{figure*}[!ht]
\centering
  \includegraphics[width=\linewidth]{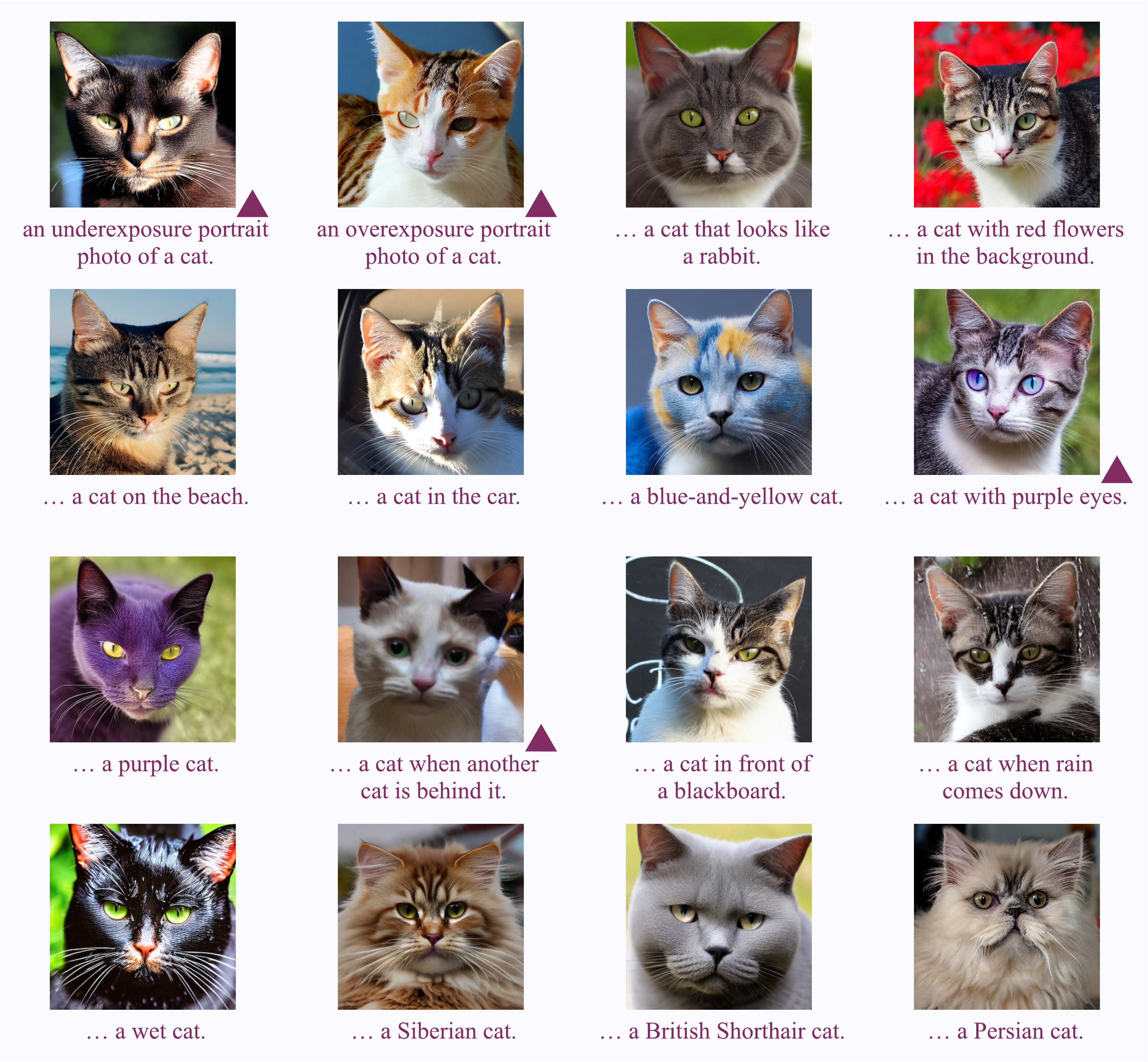}
\caption{\label{fig:cat3} Samples for text-to-image synthesis in the AFHQ Cat domain, using our \ourmethod. Notations follow Figure~\ref{fig:cat1}.
}
\end{figure*}

\subsection{Additional Samples for Domain-Specific Text-to-Image Modeling}

Figure~\ref{fig:cat1}, Figure~\ref{fig:cat2}, and Figure~\ref{fig:cat3} present more samples on the AFHQ Cat domain. Limited by the file size (50 MB), \acceptedtext{we provide more samples on the AFHQ Dog domain and the FFHQ domain in our code release. }\anonymoustext{we will provide more samples on the AFHQ Dog domain and the FFHQ domain in the supplementary materials. } 

\clearpage

\subsection{Error Analysis} 
\label{subapp:error-bias-analysis}

Our text prompts cover objects, poses, environment, attributes, lighting, and sub-population. Among them, we find some failure cases produced by our \ourmethod. In terms of objects, our method sometimes fails to model ``with AR glasses'' and ``while another cat is behind it''. In terms of pose, as expected, it does not model ambiguous descriptions consistently, such as ``looking left'' and ``looking right''. In terms of environment, unexpectedly, it fails to model ``with storm''. In terms of attributes, it sometimes fails to model low-density attributes such as ``closed eyes'', ``red eyes'', ``purple eyes'', and ``asymmetric face''. In terms of lighting, it fails to model ``underexposure'' and ``overexposure''. We also observe some unexpected biases. For instance, ``wet cats'' tend to be black; ``cats in the dark'' tend to be gray; ``underexposure cats'' tend to be black instead of being underexposure; ``overexposure cats'' tend to be white instead of being overexposure. 

\begin{table}[!ht]
    \centering
    \begin{adjustbox}{width=\linewidth}
    \begin{tabular}{@{}p{6.4cm}p{6.4cm}p{6.4cm}@{}}
        \toprule
        Prompt $\boldy$ & Positive prompt $\boldy^{+}$ & Baseline prompt $\boldy^{-}$ \\
        \midrule
        an oil painting of a couple in formal evening wear going home get caught in a heavy downpour with \textbf{no umbrellas}. & an oil painting of a couple in formal evening wear going home get caught in a heavy downpour with \textbf{no umbrellas}. & an oil painting of a couple in formal evening wear going home get caught in a heavy downpour with \textbf{an umbrella}. \\
        \midrule
        a photo of a professor \textbf{without} eyeglasses. & a photo of a professor \textbf{without} eyeglasses. & a photo of a professor \textbf{with} eyeglasses. \\
        \midrule
        a \textbf{colored} photo of a man \textbf{without} beard. & a \textbf{colored} photo of a man \textbf{without} beard. & a \textbf{black-and-white} photo of a man \textbf{with} beard. \\
        \midrule
        a bottle is in water. & a bottle is in water. & water is in a bottle. \\
        \bottomrule
    \end{tabular}
    \end{adjustbox}\smallskip
\caption{\label{tab:hard-prompt-examples} Text prompts for generalists guiding themselves to mitigate contradiction (Section~\ref{subsec:guide-generalist}). See image samples in Figure~\ref{fig:hard-experiment}. }
\end{table}

\begin{table}[!ht]
    \centering
    \begin{adjustbox}{width=\linewidth}
    \begin{tabular}{@{}p{6.4cm}p{6.4cm}p{6.4cm}@{}}
        \toprule
        Prompt $\boldy$ & Positive prompt $\boldy^{+}$ & Baseline prompt $\boldy^{-}$ \\
        \midrule
        a portrait photo of a fox. & a portrait photo of a \textbf{cute} fox. & a portrait photo of a fox. \\
        \midrule
        an oil painting of a couple in formal evening wear going home. & an oil painting of a couple in formal evening wear going home \textbf{in a heavy downpour}. & an oil painting of a couple in formal evening wear going home \textbf{in a sunny day}. \\
        \midrule
        a photo of a valley. & a \textbf{beautiful} photo of a valley. & a \textbf{not beautiful} photo of a valley. \\
        \midrule
        a photo of a waterfall. & a \textbf{beautiful} photo of a waterfall. & a \textbf{not beautiful} photo of a waterfall. \\
        \bottomrule
    \end{tabular}
    \end{adjustbox}\smallskip
\caption{\label{tab:rig-prompt-examples} Text prompts for generalists guiding themselves for continuous, rig-like control (Section~\ref{subsec:guide-generalist}). See image samples in Figure~\ref{fig:rig-experiment}. }
\end{table}

\section{Guiding Generalists (Section~\ref{subsec:guide-generalist}): Experimental Details}
\label{app:guide-generalist}

\subsection{Task Input}
\label{subapp:task-input-guide-generalist}
\ourmethod allows generalists to guide themselves (1) to improve challenging cases of text-to-image synthesis and (2) to enable continuous, rig-like controls. The input is a tuple $(\boldy, \boldy^{+}, \boldy^{-})$. For (1), the tuples for the images in Figure~\ref{fig:hard-experiment} are shown in Table~\ref{tab:hard-prompt-examples}. For (2), the tuples for the images in Figure~\ref{fig:rig-experiment} are shown in Table~\ref{tab:rig-prompt-examples}. 

\subsection{Method Details}

\noindent\textbf{Generalist Only: } Following the notations in the main text, \textit{Generalist Only} means sampling from the reverse-time SDE:
\begin{equation}
\begin{split}
    &\mathrm{d}\boldx = \bigg[\bm{f}(\boldx, t) - g(t)^{2} \guidedcolor{\tilde{\estscoreguide}(\boldx, t, \boldy)} \bigg]\mathrm{d}t + g(t)\mathrm{d}\bar{\boldw}.
\end{split}
\end{equation}
where $\guidedcolor{\tilde{\estscoreguide}(\boldx, t, \boldy)}$ means the score function after classifier-free guidance \cite{ho2021classifierfree}. It means that \textit{Generalist Only} samples: 
\begin{equation}
\begin{split}
    &\mathrm{d}\boldx = \bigg[\bm{f}(\boldx, t) - g(t)^{2} \Big(\tau_t\estscoreguide(\boldx, t, \boldy) - (\tau_t - 1)\estscoreguide(\boldx, t, \varnothing) \Big) \bigg]\mathrm{d}t + g(t)\mathrm{d}\bar{\boldw},
\end{split}
\end{equation}
where $\varnothing$ is the null token that stands for unconditional sampling, and $\tau_t$ is a hyperparameter indicating the strength of guidance. We encourage readers to read \cite{ho2021classifierfree} for more details of classifier-free guidance. 

\noindent\textbf{\titleourmethod: } When using \ourmethod to guide the generalist, we sample from the reverse-time SDE:
\begin{equation}
\begin{split}
    &\mathrm{d}\boldx = \bigg[\bm{f}(\boldx, t) - g(t)^{2} \Big(\guidedcolor{\tilde{\estscoreguide}(\boldx, t, \boldy)} + \contrastcolor{\lambda_t \big( \estscoreguide(\boldx, t, \boldy^{+}) - \estscoreguide(\boldx, t, \boldy^{-}) \big)}\Big) \bigg]\mathrm{d}t + g(t)\mathrm{d}\bar{\boldw}.
\end{split}
\end{equation}
where $\guidedcolor{\tilde{\estscoreguide}(\boldx, t, \boldy)}$ means the score function after classifier-free guidance \cite{ho2021classifierfree}. It means that \ourmethod samples from the following reverse-time SDE:
\begin{equation}
\begin{split}
    &\mathrm{d}\boldx = \bigg[\bm{f}(\boldx, t) - g(t)^{2} \Big(\tau_t\estscoreguide(\boldx, t, \boldy) - (\tau_t - 1)\estscoreguide(\boldx, t, \varnothing) + \contrastcolor{\lambda_t \big( \estscoreguide(\boldx, t, \boldy^{+}) - \estscoreguide(\boldx, t, \boldy^{-}) \big)}\Big) \bigg]\mathrm{d}t + g(t)\mathrm{d}\bar{\boldw}.
\end{split}
\end{equation}

\section{Guiding Image Editors (Section~\ref{subsec:cycle-experiment}): Experimental Details}
\label{app:cycle}

For all methods, we follow the setting described in \cite{unify} to guarantee that the best hyperparameter combination is chosen for each method, instead of using a predefined hyperparameter combination. Specifically, for each test sample, we allow each method to enumerate some combinations of hyperparameters, each combination with multiple random trials (detailed below). To select the best hyperparameter combination for each sample, we used the directional CLIP score $\mathcal{S}_{\text{D-CLIP}}$ as the criterion (higher is better). For each method, we used the DDIM sampler ($\eta = 0.1$) with 100 steps.

\noindent\textbf{SDEdit: } We enumerated the classifier-free guidance of the decoding step as $\{1, 1.5, 2, 3, 4, 5\}$; we enumerated the encoding step as $\{15, 20, 25, 30, 40, 50\}$; we ran $15$ trials for each combination. 

\noindent\textbf{\cyclediff: } We set the classifier-free guidance of the encoding step as $1$; we enumerated the classifier-free guidance of the decoding step as $\{1, 1.5, 2, 3, 4, 5\}$; we enumerated the encoding step as $\{15, 20, 25, 30, 40, 50\}$; we ran $15$ trials for each hyperparameter combination. 

\noindent\textbf{\titleourmethod for SDEdit: } Since \ourmethod helps strengthen the indented editing, we used fewer combinations to enumerate. Specifically, we enumerated the classifier-free guidance of the decoding step as $\{1, 1.5, 2, 3\}$; we enumerated the encoding step as $\{30, 40, 50\}$; we ran $15$ trials for each combination. We set $\lambda_t = 10$.  

\noindent\textbf{\titleourmethod for \cyclediff: } Since \ourmethod helps strengthen the indented editing, we used fewer combinations to enumerate. We set the classifier-free guidance of the encoding step as $1$; we enumerated the classifier-free guidance of the decoding step as $\{1, 1.5, 2, 3\}$; we enumerated the encoding step as $\{30, 40, 50\}$; we ran $15$ trials for each hyperparameter combination. We set $\lambda_t = 6$. 

}


\end{document}